\documentclass{llncs}

\usepackage[english]{babel}
\usepackage[utf8]{inputenc}

\usepackage{color,xcolor,ucs}

\usepackage{subfig}
\usepackage{floatrow}
\usepackage{tabularx}
\usepackage{float}
\usepackage{amsfonts}
\usepackage{helvet}         
\usepackage{courier}        
\usepackage{type1cm}        
\usepackage{amsmath}
\usepackage{amssymb}
\usepackage{makeidx}         
\usepackage{comment}         
\usepackage{graphicx}        
\usepackage{multicol}        
\usepackage[bottom]{footmisc}
\usepackage{bm}

\usepackage{cite}
\usepackage{url}
\urldef{\mailsa}\path|{shashank,antony.thomas, asaff,vadim.indelman}@technion.ac.il|

\usepackage[unicode=true, bookmarks=true,colorlinks=true]{hyperref}
\usepackage{xr-hyper}

\usepackage{algpseudocode,algorithm,algorithmicx}

\usepackage{xspace}
\usepackage{rotating}

\usepackage{tikz}
\usepackage{tkz-graph}
\usetikzlibrary{arrows,shapes,shadows,positioning,calc}

\usepackage{graphicx}
\usepackage{threeparttable}
\usepackage{multirow}
\usepackage[font=scriptsize,labelfont=bf]{caption}

\DeclareMathOperator*{\argmin}{arg\,min}

\newcommand{\alias}[1]{\ensuremath{\{{#1}\}_{\textbf{aliased}}}}

\newcommand{\prob}[1]{\ensuremath{\mathbb{P}({#1})}}

\newcommand*\Let[2]{\State #1 $\gets$ #2}
\algrenewcommand\algorithmicrequire{\textbf{Input:}}
\algrenewcommand\algorithmicensure{\textbf{Input:}}
\algnewcommand{\LineComment}[1]{\State \(\triangleright\) #1}

\newcommand{\event}[1]{\ensuremath{A_{#1}}\xspace}

\newcommand{\events}{\ensuremath{\{\event{\mathbb{N}}\}}\xspace}

\newcommand{\his}{\ensuremath{{\cal H}}\xspace}

\newcommand{\das}{\texttt{DAS}}

\title{Robust Active Perception via Data-association aware Belief Space Planning}

\author{Shashank Pathak%
	\and Antony Thomas\and Asaf Feniger \and Vadim Indelman}
\institute{Department of Aerospace Engineering,\\
	Technion - Israel Institute of Technology\\
	\mailsa\\
	\url{http://vindelman.net.technion.ac.il}}

\date{}

\begin{document}

\maketitle

\begin{abstract}
	We develop a belief space planning (BSP) approach that advances the state of the art by incorporating reasoning about data association (DA) within planning, while considering additional sources of uncertainty. Existing BSP approaches typically assume data association is given and perfect, an assumption that can be harder to justify while operating, in presence of localization uncertainty, in ambiguous and perceptually aliased environments. In contrast, our data association aware belief space planning (DA-BSP) approach explicitly reasons about DA within belief evolution, and as such can better accommodate these challenging real world scenarios. In particular, we show that due to perceptual aliasing, the posterior belief becomes a mixture of probability distribution functions, and design cost functions that measure the expected level of ambiguity and posterior uncertainty. Using these and standard costs (e.g.~control penalty, distance to goal) within the objective function, yields a general framework that reliably represents action impact, and in particular, capable of active disambiguation. Our approach is thus applicable to robust active perception and autonomous navigation in perceptually aliased environments. We demonstrate key aspects in basic and realistic simulations.

\end{abstract}

\section{Introduction}
\label{sec:introduction}
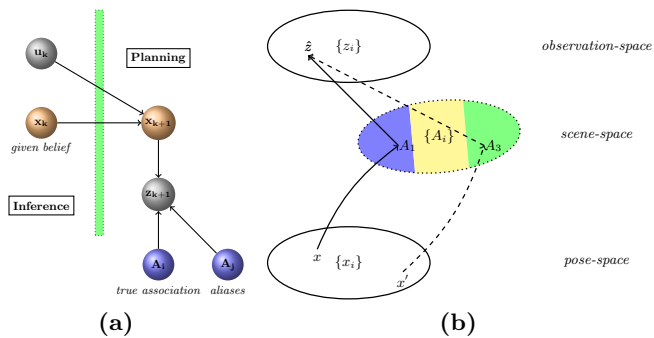
\begin{figure}
	\centering
	\subfloat[]{
	\scalebox{0.5}{
	\GraphInit[vstyle = Shade]
  
\begin{tikzpicture}
\tikzset{Poses/.style = {shape = circle,
ball color = orange!50,
text = black,
font = \bfseries,
inner sep = 2pt,
outer sep = 0pt,
minimum size = 24 pt},
Observations/.style = {shape = circle,
ball color = gray!50,
text = black,
font = \bfseries,
inner sep = 2pt,
outer sep = 0pt,
minimum size = 24 pt},
Objects/.style = {shape = circle,
ball color = blue!50,
text = black,
font = \bfseries,
inner sep = 2pt,
outer sep = 0pt,
minimum size = 24 pt}}
\tikzset{
  Gamma/.style = { rectangle, rounded corners, draw,
                        minimum width = 2em, fill = yellow!50,
                        text = red, font = \bfseries },
  Constraint/.append style = {thick,->, bend right} ,
  Timestep/.append style = {thick, dotted,->},
  Observe/.append style = {thick, ->},
  Msg/.style = { rectangle,draw,thick,minimum width=0.1cm,minimum height=0.5cm, font = \bfseries },
  Cmnt/.style = { rectangle, minimum width=0.1cm,minimum height=0.5cm }
  }

\SetGraphUnit{2.5}
 \node[Observations](uk){$\mathbf{u_k}$};
  \node[Poses,below=of uk](Xk){$\mathbf{x_k}$};
  \node[rectangle,draw,dotted,minimum height=6cm,fill=green!50,right=of Xk](Prt){};
  \node[Poses,right=of Prt](Xk1){$\mathbf{x_{k+1}}$};
  \node[Observations, below=of Xk1](Zk1){$\mathbf{z_{k+1}}$};

  \node[Objects, below=of Zk1](Ai){$\mathbf{A_i}$};
  \node[Objects, right=of Ai](Aj){$\mathbf{A_j}$};

  \draw[Observe](Xk) to node {} (Xk1);
  \draw[Observe](uk) to node {} (Xk1);
  
  
  \draw[Observe](Xk1) to node {} (Zk1);
  \draw[Observe](Ai) to node {} (Zk1);
  \draw[Observe](Aj) to node {} (Zk1);


  \node[Cmnt, below=of Xk,yshift=1cm](AsmDA){\textit{given belief}};
  \node[Cmnt,below=of Ai,yshift=1cm](TrAq){\textit{true association}};
  \node[Cmnt,below=of Aj,yshift=1cm](Als){\textit{aliases}};
  
  \node[Msg,below=of AsmDA](Inf){\textbf{Inference}};
  \node[Msg,above=of Xk1](Pln){\textbf{Planning}};

\end{tikzpicture}}}%
	\subfloat[]{
	\scalebox{0.6}{
		\begin{tikzpicture}

\def\elpA{1.8 }
\def\elpB{0.8}
\def\sceneX{2.2}
\def\sceneY{4}
\def\sceneRot{5}

\def\regionOne{({\sceneX - \elpA},{\sceneY-\elpB}) rectangle ({\sceneX-\elpA/3},{\sceneY+\elpB})};
\def\regionTwo{({\sceneX - \elpA/3},{\sceneY-\elpB}) rectangle ({\sceneX+\elpA/3},{\sceneY+\elpB})};
\def\regionThree{({\sceneX + \elpA/3},{\sceneY-\elpB}) rectangle ({\sceneX+\elpA},{\sceneY+\elpB})};

\def\spaceX{({\sceneX-2},{\sceneY-2.8}) ellipse ({\elpA} and {\elpB})};
\def\spaceA{({\sceneX},{\sceneY}) ellipse ({\elpA} and {\elpB})};
\def\spaceZ{({\sceneX-2},{\sceneY+2}) ellipse ({\elpA} and {\elpB})};

\draw[thick] \spaceX;
\node at ({\sceneX-2},{\sceneY-2.8}) {$\{x_i\}$};
\draw[thick] \spaceZ;
\node at ({\sceneX-2},{\sceneY+2}) {$\{z_i\}$};

\begin{scope}[rotate around={{\sceneRot}:({\sceneX},{\sceneY})}]
\begin{scope}
\clip \regionOne;
\fill[blue!50] \spaceA;
\end{scope}
\begin{scope}
\clip \regionTwo;
\fill[yellow!50] \spaceA;
\end{scope}
\begin{scope}
\clip \regionThree;
\fill[green!50] \spaceA;
\end{scope}
\draw[thick,dotted] \spaceA;
\end{scope}
\node at ({\sceneX},{\sceneY}) {$\{A_i\}$};

\draw[thick,->] ({\sceneX-2.7},{\sceneY-2.5}) to [bend left=15] ({\sceneX-0.9},{\sceneY-0.2});
\draw[thick,->] ({\sceneX-0.9},{\sceneY-0.2}) to ({\sceneX-2.9},{\sceneY+1.8});
\node at ({\sceneX-2.9},{\sceneY+2}) {$\hat z$};
\node at ({\sceneX-0.7},{\sceneY-0.2}) {$A_1$};
\node at ({\sceneX-2.7},{\sceneY-2.7}) {$x$};

\draw[thick,dashed,->] ({\sceneX-0.8},{\sceneY-3}) to [bend right=15] ({\sceneX + \elpA/3+.4},{\sceneY-0.2});
\draw[thick,dashed,->] ({\sceneX + \elpA/3+.4},{\sceneY-0.2}) to ({\sceneX-2.9},{\sceneY+1.8});
\node at ({\sceneX-2.9},{\sceneY+2}) {$\hat z$};
\node at ({\sceneX + \elpA/3+.6},{\sceneY-0.2}) {$A_3$};
\node at ({\sceneX-0.8},{\sceneY-3.2}) {$x'$};

\node at ({\sceneX+3.5},{\sceneY-2.8}) {\textit{pose-space}};
\node at ({\sceneX+3.5},{\sceneY}) {\textit{scene-space}};
\node at ({\sceneX+3.5},{\sceneY+2}) {\textit{observation-space}};

\end{tikzpicture}}\label{fig:GenGraphModel}}
	\caption{\scriptsize (a) Generative graphical model. Standard BSP approaches assume data association (DA) is given and perfect (\das{}). We incorporate data association aspects within BSP and thus can reason about ambiguity (e.g.~perceptual aliasing) at a decision-making level. (b) Schematic representation of pose, scene and observation spaces. Scenes $A_1$ and $A_3$ when viewed from perspective $x$ and $x'$ respectively, produce the same nominal observation $\hat z$, giving rise to \emph{perceptual aliasing}. }
	\label{fig:schema_spaces}
	\vspace{-7pt}
\end{figure}

In the context of partially observable Markovian systems, planning over belief space (BSP) under some simplifying assumptions, provides scalable applications including autonomous navigation, object grasping and manipulation, active SLAM, and robotic surgery. In presence of uncertainty, such as in robot motion and sensing, the true state of variables of interest (e.g. robot poses), is unknown and can only be represented by a probability distribution over possible states, given available data. 
This distribution, the belief space, is inferred using probabilistic approaches based on incoming sensor observations and prior knowledge. The corresponding problem is an instantiation of a partially observable Markov decision problem (POMDP) \cite{Kaelbling98ai}. Apart from simplifying structural assumptions -- such as Gaussian noise around a given observation and motion model -- state-of-the-art BSP approaches typically assume data association to be given and perfect (see Figure \ref{fig:GenGraphModel}), i.e.~the robot is assumed to correctly perceive the environment to be observed by its sensors, given a candidate action. For brevity, we shall call it \das{}. In reality, the world is often full of ambiguity, that together with other sources of uncertainty, make perception a challenging task. As an example, matching images from two different but similar in appearance places, or attempting to recognise an object that is similar in appearance, from the current viewpoint, to another object. Both cases are examples of ambiguous situations, where na\"{i}ve and straightforward approaches using \das{} are likely to yield incorrect results, i.e. mistakenly considering the two places as same, and incorrectly associating the observed object.

Thus, in presence of ambiguity, \das{} may lead to incorrect posterior beliefs and as a result, to sub-optimal actions. More advanced approaches are therefore required to enable reliable operation in ambiguous conditions, approaches often referred to as (active) robust perception. These approaches typically involve probabilistic data association and hypothesis tracking given available data. Thus, for the object detection example, each hypothesis may represent a candidate object from a given database that the current observation (e.g. image or point-cloud) is successfully registered to. Similarly, one might reason probabilistically regarding perceptual aliasing, as in the first example above, which would also involve probabilistic data association. Yet, existing robust perception approaches focus on the passive case, where robot actions are externally determined and given, while the closely related approaches for active object detection and classification consider the robot to be perfectly localised.

In this work we develop a general data association aware belief space planning (DA-BSP) framework capable of better handling complexities arising in real world, possibly perceptually aliased, scenarios. We rigorously incorporate reasoning about data association within belief space planning, while also considering other sources of uncertainty (motion, sensing and environment). In particular, we show our framework can be used for active disambiguation by determining appropriate actions, e.g. future viewpoints, for increasing confidence in a certain data association hypothesis.

\emph{Organization of the paper}: After discussing related work and stating our contributions, we formulate the considered problem in Section \ref{sec:problem-formulation}. In Section \ref{sec:concept} we provide concept overview and then discuss in detail the proposed approach, while demonstrating key aspects in simulated basic and realistic scenarios in Section \ref{sec:results}. Finally, in Section \ref{sec:conclusions} we conclude the discussion and suggest potential directions for future research. 

\subsection{Related Work}
\label{sec:related-work}

Calculating optimal solutions to POMDP is computationally intractable (PSPACE-complete) \cite{Papadimitriou87math} for all but the smallest problems. The vast research area of approximate approaches (with reduced computational complexity) can be roughly segmented into  point-based value iteration methods \cite{Pineau06jair, Kurniawati08rss}, simulation based \cite{Stachniss05rss} and sampling based approaches \cite{Prentice09ijrr, Bry11icra, AghaMohammadi14ijrr}, and direct trajectory optimization \cite{VanDenBerg12ijrr, Patil14wafr, Indelman15ijrr} methods. In all cases, finding the (locally) optimal actions involves evaluating a given objective function while considering future observations to be acquired as a result of each candidate action. They all assume \das{}. For example, it is typically assumed that the robot can be localised by making observations of known landmarks or beacons (see, e.g.~\cite{Prentice09ijrr, AghaMohammadi14ijrr}), while assuming to correctly associate each future measurement with an appropriate landmark. Though reasonable in certain scenarios, \das{} becomes unrealistic in the presence of perceptually aliased environments (two scenes that look alike) and localisation uncertainty, as in this work.

The issue of perceptual aliasing has been considered in the earlier works on POMDP planning, though again with highly simplified scenarios, since the data-association further complicates the problem. In a slightly separate line of research, the approaches that study the issue were in the context of  multiple hypothesis tracking(see~\cite{Reid79itac} for earliest work on MHT) or more recently, of active robust perception. Both these approaches rely on passive and often non-parametric approaches, through various filtering techniques; we refer an interested reader to the book~\cite{Jazwinsky70} and tutorial~\cite{Arulampalam02} for further details. For example, \cite{Vo06tsp} proposed using Gaussian mixture probability hypothesis density (PHD) filter. To the best of our knowledge, such approaches are not considered in the context of active planning.

Coming back to scalable planning methods such as BSP, we note that while the traditional BSP approaches had typically assumed the environment to be accurately known (e.g.~a given map), recent works, including  \cite{Chaves14iros, Chaves15iros, Walls15iros, Kim14ijrr, Indelman15ijrr}, relax this assumption and model the uncertainty of the environment mapped thus far within the belief. The corresponding framework is thus tightly related to active SLAM, with the well known trade-off between exploration and exploitation. Recent work \cite{Kim14ijrr, Indelman15ijrr, Chaves15iros, Walls15iros} in this branch focused in particular on probabilistically modelling what future observations will be obtained given a candidate action. Though none of them relax \das{} assumption.

In the last few years, the SLAM research community has investigated approaches to be resilient to false data association (outliers) overlooked by front-end algorithms (e.g.~image matching), see e.g.~\cite{Sunderhauf12icra, Olson13ijrr, Carlone14iros, Indelman14icra, Indelman16csm}. However these approaches, also known as robust graph optimization approaches, are developed only for the passive problem setting, i.e.~robot actions are given and externally determined. In contrast, we consider a complimentary \emph{active} framework that incorporates  data association aspects within BSP.

Our approach is also tightly related with recent work on active hypothesis disambiguation in the context object detection and classification \cite{Atanasov14tro, Sankaran15arxiv, Lauri15rssws, Wong15ijrr, Patten16ral}. Given hypotheses regarding object class and pose, these approaches aim to find a sequence future viewpoints that will lead to disambiguation, i.e.~identifying the correct hypothesis. However, these approaches assume the sensor is perfectly localized and can be shown to be a specific case of DA-BSP.

Probably the closest work to our approach is by Agarwal et al.~\cite{Agarwal15arxiv}, where the authors also consider hypotheses due to ambiguous data association and develop a BSP approach for active disambiguation. However, unlike them, DA-BSP considers ambiguous data association also in posterior and thus does not require a guarantee of fully disambiguating action in the future.

\subsection{Contributions}

To summarize, our main contributions in this paper\footnote{Earlier versions of this paper appeared in \cite{Pathak16icra_ws} and \cite{Pathak16ecai}.} are as follows: (a) relaxing the data-association-is-solved assumption for a general data-association aware BSP framework (DA-BSP) with GMM priors (b) considering active data-association aspect for both planning and inference, hence providing a closed-loop framework (c) reducing some of the known recent BSP approaches to a degenerate cases of DA-BSP (d) demonstrating empirical results in support of two claims: data-association is crucial for a robust BSP and the principled approach of DA-BSP can be scalable enough to be applied on practical problems.

\section{Notations and Problem Formulation}
\label{sec:problem-formulation}

Consider a robot operating in a partially known or pre-mapped environment which can be ambiguous and perceptually aliased. The robot takes observations of different scenes and objects in the environment, and uses these observations to infer application-dependent random variables of interest (e.g.~past and current robot poses).  The following three spaces are involved in the considered problem, as shown in Figure \ref{fig:schema_spaces}: \emph{pose-space}, \emph{scene-space} and \emph{observation-space}. 

\emph{Pose-space} involves all possible perspectives a robot can take with respect to a given environment model and in the context of task at hand. We denote the robot pose at time step $k$ by $x_k$ and a sequence of poses from $0$ up to $k$ by $X_k\doteq \{x_0,\ldots,x_k\}$. Given all controls $u_{0:k-1}\doteq \{u_0,\ldots,u_{k-1}\}$ and observations $Z_{0:k}\doteq \{Z_0,\ldots,Z_k\}$ up to time step $k$, the posterior probability distribution function\footnote{Strictly speaking, this is either the probability mass function or the probability density function for a discrete or a continuous random variable, respectively.} is defined as $\prob{X_k|u_{0:k-1},Z_{0:k}}$. For notational convenience, we define below  histories $\mathcal{H}_k$ and $\mathcal{H}_{k+1}^-$ and rewrite the posterior pdf (\emph{belief}), at time $k$ as $b[X_k] \doteq \prob{X_k|\mathcal{H}_k}$.
\begin{eqnarray}
\mathcal{H}_k\doteq \{u_{0:k-1},Z_{0:k}\} \ \ , \ \ \mathcal{H}_{k+1}^- \doteq \mathcal{H}_k \cup \{u_k\}.
\label{eq:Histories}
\end{eqnarray}
The \emph{scene-space} involves a discrete set of objects or scenes, denoted by the set \events, in the given world model, and which can be detected through the sensors of the robot. We will use symbols \event{i} and \event{j} to denote such typical scenes. Note that even if the objects are identical, they are distinct in scene space. This will be important when we shall consider the cases where the objects look similar from some perspectives. Finally, \emph{observation-space} is the set of all possible observations that the robot is capable of obtaining when considering its mission and sensory capabilities. 

We consider probabilistic motion and observation models 
\begin{eqnarray}
x_{k+1} = f(x_k,u_k)+w_k \ \ , \ \ z_{k} = h(x_k,A_i)+v_k,
\label{eq:MotionObsModel}
\end{eqnarray}
and denote them  by $\prob{x_{k+1}|x_k,u_k}$ and $\prob{z_k|x_k,A_i}$, respectively. As common in literature, we consider Gaussian zero-mean process and measurement noise $w_i \sim \mathcal{N}(0,\Sigma_w)$ and $v_k \sim \mathcal{N}(0,\Sigma_v)$, with known noise covariance matrices $\Sigma_w$ and $\Sigma_v$. Here, $h(x_k, \event{i})$ is a noise-free observation which we would refer as \emph{nominal} or \emph{predicted} observation $\hat z$, that corresponds to observing scene $A_i$ from pose $x_k$.

Given a prior  $\prob{x_0}$ and motion and observation models (\ref{eq:MotionObsModel}), the joint posterior pdf at the current time $k$ can be written as
\begin{eqnarray}
\prob{X_k|\mathcal{H}_k} = \prob{x_0}\prod_{i=1}^k \prob{x_i|x_{i-1},u_{i-1}} \prob{Z_i|x_i,A_{i}}.
\label{eq:posterior_k}
\end{eqnarray} 
Note that \das{} is the underlying assumption in the above equation.

If the prior $\prob{x_0}$ is Gaussian, it is not difficult to show that $b[X_k]$ is also a Gaussian with some mean $\hat{X}_k$ and covariance $\Sigma_k$ that can be efficiently calculated via maximum a posteriori (MAP) inference, see e.g.~\cite{Kaess12ijrr}. It is also valid in case where the environment model is given but uncertain, and when this model is unknown a priori and instead is constructed on-line within SLAM framework. However, in this paper we consider a more general case where the prior belief is modeled by a Gaussian mixture model (GMM). Such a situation can arise, for example, in the kidnapped robot problem in a perceptually aliased environment (e.g. different similar in appearance rooms), where matching sensor observations against a given map would indicate several most probable robot locations. In such a case the belief at time $k$ can be represented by a GMM,
\begin{eqnarray}
b[X_k] = \sum_{j=1}^{M_{k}} \xi_{k}^j \prob{X_{k}|\mathcal{H}_{k}, \gamma=j}, 
\label{eq:GMM_k}
\end{eqnarray}
where $M_{k}$ is the number of components (or modes), the $j$th component is represented by the weight $\xi_{k}^j \doteq \prob{\gamma=j|\mathcal{H}_{k}}$, modeling the probability of the robot being in that component,  and by the conditional Gaussian 
\begin{equation}
b[X_{k}^j]\doteq \prob{X_{k}|\mathcal{H}_{k}, \gamma=j}=\mathcal{N}(\hat{X}_{k}^j,\Sigma_{k}^j),
\end{equation}
with appropriate mean $\hat{X}_{k}^j$ and covariance $\Sigma_{k}^j$. Here, $\gamma$ is an indicator variable denoting the component number.

Given the belief at time $k$, one can reason about the robot's best future actions that would minimize (or maximize) an objective function $J$. %
\begin{equation}
J(u_{k}) = \mathbb{E}\left\{c \left(
b[X_{k+1}], u_{k} \right) \right\},
\label{eq:ObjFunction}
\end{equation}
where the expectation is over the (unknown) future observation $z_{k+1}$, and $c(.)$ is  the immediate cost.

The posterior belief at time $t_{k+1}$ is a function of control $u_{k}$ and observation $z_{k+1}$, i.e.
\begin{equation}
b[X_{k+1}] \doteq \prob{X_{k+1}|\mathcal{H}_{k+1}} \equiv \mathbb{P}(X_{k+1}| \mathcal{H}_{k}, z_{k+1}, u_{k}).
\label{eq:Belief_l}
\end{equation}
Note that, according to Eq.~(\ref{eq:ObjFunction}),  we need to calculate the posterior belief (\ref{eq:Belief_l}) for \emph{each} possible value of $z_{k+1}$. 

Similarly, we define the propagated joint belief as
\begin{eqnarray}
b[X_{k+1}^-]\doteq \prob{X_{k}|\mathcal{H}_{k}} \prob{x_{k+1}|x_{k},u_{k}},
\label{eq:PredictedBelief}
\end{eqnarray}
from which the marginal belief over the future pose $x_{k+1}$ can be calculated as $b[x_{k+1}^-] \doteq \int_{\lnot x_{k+1}} b[X_{k+1}^-]$. 

In particular, the propagated belief at the first look ahead step, given the GMM belief (\ref{eq:GMM_k}) at time $k$ is
\begin{eqnarray}
b[X_{k+1}^-] =  \sum_{j=1}^{M_{k}} \xi_{k+1}^{j-} b[X_{k+1}^{j-}], 
\label{eq:PredictedBelief_step1GMM}
\end{eqnarray}
with $\xi_{k+1}^{j-} \doteq \prob{\gamma_{k+1}=j|\mathcal{H}_{k+1}^-}\equiv \xi_{k}^j$, and $b[X_{k+1}^{j-}] \doteq \prob{X_{k+1}|\mathcal{H}_{k+1}^-, \gamma_{k}=j}= b[X_{k}^j] \prob{x_{k+1}|x_k,u_k}$.

As earlier, with \das{} assumption, one can consider for each specific value of $z_{k+l}$ the corresponding observed scene $A_i$, and express the posterior (\ref{eq:Belief_l}) recursively as
\begin{equation}
b[X_{k+1}] \!=\! \eta b[X_{k+1}^-] \prob{z_{k+1}|x_{k+1},A_i},
\label{eq:Posterior_givenAi} 
\end{equation}
which can be represented as $b[X_{k+1}]=\mathcal{N}(\hat{X}_{k+1}, \Sigma_{k+1})$ with appropriate mean $\hat{X}_{k+1}$ and covariance $\Sigma_{k+1}$. The optimal control is then defined as:\\
 $u_{k}^{\star} \doteq \argmin_{u_{k}} J(u_{k})$.

\das{} assumption simplifies greatly the above formulation. Yet, in practice, determining data association reliably is often a non trivial task by itself, especially when operating in perceptually aliased environments. An incorrect data association (wrong scene $A_i$ in Eq.~(\ref{eq:Posterior_givenAi})) can lead to catastrophic results, see, e.g.~\cite{Indelman14icra, Indelman14rss_ws, Indelman16csm}. In this work we relax this restricting assumption and rigorously incorporate data association aspects within belief space planning and inference considering the underlying distributions are GMMs.

\section{Approach}
\label{sec:concept}

Given some candidate action (or sequence of actions) and the belief at planning time $k$, we can reason about a future observation $z_{k+1}$ (e.g.~an image) to be obtained once this action is executed; its actual value is unknown. All the possible values such an observation can assume should be taken into account while evaluating the objective function; hence, the expectation operator in Eq.~(\ref{eq:ObjFunction}). When written explicitly it transforms to
\begin{equation}
J(u_k) \!\! \doteq  \!\! \int_{z_{k+1}} \!\!\!\!\!\!\!  \overbrace{\prob{z_{k+1} \mid \mathcal{H}_{k+1}^-}}^{(a)}\;\;  \!\! c\left(\overbrace{ \prob{X_{k+1}|\mathcal{H}_{k+1}^-,z_{k+1}} }^{(b)} \! \right) 
\label{eq:cost_function}
\end{equation}
The two terms $(a)$ and $(b)$ in the above equation have intuitive meaning: for each considered value of $z_{k+1}$, $(a)$ represents how likely is it to get such an observation when both the history \his and control $u_k$ are known, while $(b)$ corresponds to the posterior belief \emph{given} this specific  $z_{k+1}$.

Considering \das{} means we can correctly associate each possible measurement $z_{k+1}$ with the corresponding scene $A_i$ it captures, as in Eq.~(\ref{eq:Posterior_givenAi}). Yet, it is unknown from what future robot pose $x_{k+1}$ the actual observation $z_{k+1}$ will be acquired, since the \emph{actual} robot pose $x_k$ at time $k$ is unknown and the control is stochastic. Indeed, as a result of action $u_k$, the robot actual (true) pose $x_{k+1}$ can be anywhere within the propagated belief $b[x_{k+1}^-]$. In inference, we have a similar situation with the key difference that the observation $z$ has been acquired. We must first associate the captured measurement $z$ with the scene or object $A_i$ it describes, i.e.~write the appropriate measurement likelihood term in the posterior (\ref{eq:posterior_k}). 

In BSP framework, solved data association means that for each such observation $z\in \{z\}$ the corresponding observed scene $A_i\in \mathcal{A}$ is known. In contrast, we do not assume this, and instead reason about possible scenes or objects that the future observation $z_{k+1}$ could be generated from, see Figures~\ref{fig:GenGraphModel} and \ref{fig:schema_spaces}.

\paragraph{Parsimonious data association:} 
 Incorporating data-association is expensive. However, if the environment has only distinct scenes or objects, then for each specific value of $z_{k+1}$, there will be only one scene $A_i$ that can generate such an observation according to the model (\ref{eq:MotionObsModel}). In case of perceptually aliased environments, there could be also several scenes (or objects) that are either completely identical, or have a similar visual appearance when observed from appropriate viewpoints. They could equally well explain the considered observation $z_{k+1}$. Thus, there are several possible associations $\{A_i\}$ and due to localisation uncertainty determining which association is the correct one is not trivial. As we show in the sequel, in these cases the posterior $b[X_{k+1}]$ (term $(b)$ in Eq.~(\ref{eq:cost_function})) becomes a Gaussian mixture with appropriate weights that we rigorously compute. Additionally, the weight updates are capable of discriminating against unlikely data-associations, during the planning steps.

\paragraph{Perceptual aliasing:} 
 Intuitively speaking, perceptual aliasing occurs when an object different from the actual one, produces the same observation and thereby is an alias, in the sense of perception, to the true object. Consider two notions of perceptual aliasing: \emph{exact} and \emph{probabilistic}. Exact perceptual aliasing of  scenes \event{i} and \event{j} is defined as 
$\exists x, x',\; h(x,\event{i}) = h(x',\event{j})$, and will be denoted in this paper by $\alias{A_i,A_j}$. In other words, the same nominal (noise-free) observation $\hat{z}$ can be generated by observing different scenes, possibly from different viewpoints. Such a situation is depicted in Figure \ref{fig:schema_spaces}.
A probabilistic perceptual aliasing is a more general form of aliasing, which can be defined as $\exists x, x',\; | \prob{z | \event{i},x} - \prob{z | \event{j},x'}|<\epsilon$ for some small threshold $\epsilon$.

\subsection{Computing the term (a) : $\mathbb{P}(z_{k+1}|\mathcal{H}_{k+1}^-)$}\label{sec:TermA}
Applying total probability over non-overlapping scene space $\events$ and marginalizing over all possible robot poses, yields
\begin{equation}
\prob{z_{k+1}  |  \mathcal{H}_{k+1}^-} \!\! \equiv \!\! \sum_i^{|A_{\mathbb{N}}|} \! \int_x \! \prob{z_{k+1}, x, A_i \! \mid \! \mathcal{H}_{k+1}^-} \!\! \doteq \!\! \sum_i^{|A_{\mathbb{N}}|} \!\! w_{k+1}^i.
\label{eq:TermA_Wi}
\end{equation}
As seen from the above equation, to calculate the likelihood of obtaining some observation $z_{k+1}$, we consider separately, for each scene $A_i \in \events$, the likelihood that this observation was generated by scene $A_i$. This probability is captured for each scene $A_i$ by a corresponding weight $w_{k+1}^i$; these weights are then summed to get the actual likelihood of observation $z_{k+1}$. As will be seen below, these weights naturally account for perceptual aliasing aspects for each considered  $z_{k+1}$.

In practice, instead of considering the entire scene space \events that could be huge, the availability of the belief $b[X_{k+1}^-]$ makes it possible to consider only those scenes that could be actually observed from  viewpoints with non-negligible probability according $b[X_{k+1}^-]$, e.g.~within $3$ standard deviations of uncertainty for each GMM component. In the following, however, we proceed while reasoning about the entire scene space \events. 

Proceeding with the derivation further, using the chain rule we compute
\begin{equation}
\sum_i \int_x \prob{z_{k+1} \mid x, A_i, \mathcal{H}_{k+1}^-} \prob{A_i, x \mid \mathcal{H}_{k+1}^-}
\end{equation}
However, since $\prob{A_i, x \mid \mathcal{H}_{k+1}^-} = \prob{A_i | x \mid \mathcal{H}_{k+1}^-} b[x_{k+1}^- = x]$, we get
\begin{equation}
\sum_i^{|A_{\mathbb{N}}|} \int_x \prob{z_{k+1} | x, A_i, \mathcal{H}_{k+1}^-} \prob{A_i | \mathcal{H}_{k+1}^-, x } b[x_{k+1}^-=x].
\end{equation}

Thus,
\begin{eqnarray}
w_{k+1}^i \! \doteq \!\! \int_x \! \!\prob{z_{k+1} | x, A_i, \mathcal{H}_{k+1}^-} \prob{A_i | \mathcal{H}_{k+1}^-, x } b[x_{k+1}^-\!\!=\!x].
\label{eq:wi_def}
\end{eqnarray}
Since the propagated belief (\ref{eq:PredictedBelief_step1GMM}), from which $b[x_{k+1}^-]$ is calculated, is a GMM, we 
can replace $b[x_{k+1}^-=x]$ with $\sum_{j=1}^{M_k} \xi_{k+1,j}^- b[x_{k+1,j}^-=x]$.

Here, $\prob{z_{k+1} \mid \event{i}, x, \mathcal{H}_{k+1}^-}\equiv \prob{z_{k+1} \mid \event{i}, x}$ is the standard measurement likelihood term, while  $\prob{\event{i}\mid \mathcal{H}_{k+1}^-,x}$ represents the \emph{event likelihood}, which denotes the probability of scene $A_i$ to be observed from viewpoint $x$. In other words, this scenario-dependent term encodes from what viewpoints each scene $A_i$ is observable and could also model occlusion and additional aspects. As such, this term can be determined given a model of the environment and thus, in this work, we consider this term to be given.

The weights $w_{k+1}^i$ (\ref{eq:wi_def}) naturally capture \emph{perceptual aliasing} aspects: consider some observation $z_{k+1}$ and the corresponding generative model $z_{k+1}=h(x^{tr},A^{tr})+v$ with appropriate unknown \emph{true} robot pose $x^{tr}$ and scene $A^{tr}\in \events$. Clearly, the measurement likelihood $\prob{z_{k+1} \mid x, A_i, \mathcal{H}_{k+1}^-}$ will be high when evaluated for $A_i=A^{tr}$ and in vicinity of $x^{tr}$. Note that we will necessarily consider such a case, since according to Eq.~(\ref{eq:TermA_Wi}) we separately consider each scene $A_i$ in \events, and, given $A_i$, we reason about all poses $x$ in Eq.~(\ref{eq:wi_def}). In case of perceptual aliasing, however, there will be also another scene(s) $A_j$ which could generate the same observation $z_{k+1}$ from appropriate robot pose $x'$. Thus, the corresponding measurement likelihood term to $A_j$ will also be high for $x'$.

However, the actual value of $w_i$ (for each $A_i \in \events$) depends, in addition to the measurement likelihood, also on the mentioned-above event likelihood and on the GMM belief $b[x_{k+1}^-]$, with the latter weighting the probability of each considered robot pose $x$. This correctly captures the intuition that those observations $z$ with low-probability poses $b[x_{k+1}^-=x^{tr}]$ will be unlikely to be actually acquired, leading to low value of $w_i$ with $A_i=A^{tr}$. Since $b[x_{k+1}^-]$ is a GMM with $M_k$ components, low-probability pose $x^{tr}$ corresponds to low probabilities $b[x_{k+1}^{j-}=x^{tr}]$ for each component $j\in\{1,\ldots,M_k\}$. However, the likelihood term (\ref{eq:TermA_Wi}) could still go up in case of perceptual aliasing, where the aliased scene $A_j$ generates a similar observation to $z_{k+1}$ from viewpoint $x'$ with latter being more probable, i.e.~high probability $b[x_{k+1}^-=x']$.

In practice, calculating the integral in Eq.~(\ref{eq:wi_def}) can be done efficiently considering separately each component of the GMM $b[x_{k+1}^-]$. Each such component is a Gaussian that is multiplied by the measurement likelihood $\prob{z_{k+1} \mid \event{i}, x, \his}$ which is also a Gaussian and it is known that a product of Gaussians remains a Gaussian. The integral can then be only calculated for the window where event likelihood is non-zero i.e $\prob{\event{i} \mid x, \his} > 0$. For general probability distributions, the integral in Eq.~(\ref{eq:wi_def}) should be computed numerically. Since in practical applications $\prob{\event{i} \mid x, \his}$ is sparse w.r.t. $x$, this computational cost is not severe.

\subsection{Computing the term (b) : $\prob{X_{k+1}|\mathcal{H}_{k+1}^-,z_{k+1}}$}\label{sec:TermB}
The term $(b)$, $\prob{X_{k+1}|\mathcal{H}_{k+1}^-,z_{k+1}}$, represents the posterior probability conditioned on observation $z_{k+1}$. This term can be similarly calculated, with a key difference: since the observation $z_{k+1}$ is given, it must have been generated by \emph{one} specific (but unknown) scene $A_i$ according to  measurement model (\ref{eq:MotionObsModel}). Hence, also here, we consider all possible such scenes and weight them accordingly, with weights $\tilde{w}_{k+1}^i$ representing the probability of each scene $A_i$ to have generated the observation $z_{k+1}$. As will be seen next, the posterior $\prob{X_{k+1}|\mathcal{H}_{k+1}^-,z_{k+1}}$ is a GMM with $M_{k+1}$ components.

Applying total probability over non-overlapping \events and chain-rule, we get: 
\begin{equation}
\prob{X_{k+1}|\mathcal{H}_{k+1}^-,z_{k+1}} = \sum_{i=1}^{|A_{\mathbb{N}}|} \prob{X_{k+1} \mid \mathcal{H}_{k+1}^-, z_{k+1}, \event{i}} \cdot \prob{\event{i} \mid \mathcal{H}_{k+1}^-, z_{k+1}}.
\end{equation}

The first term, $\prob{X_{k+1} \mid \mathcal{H}_{k+1}^-, z_{k+1}, \event{i}}$, is the posterior belief conditioned on observation $z_{k+1}$, history $\mathcal{H}_{k+1}^-$, as well as a candidate scene \event{i} that supposedly generated the observation $z_{k+1}$. It is not difficult to show that this posterior is actually the GMM
\begin{equation}
 \prob{X_{k+1} \mid \mathcal{H}_{k+1}^-, z_{k+1}, \event{i}} = \sum_{j=1}^{M_k}\xi_{k}^j b[X_{k+1}^{j+}|A_i],
 \label{eq:PosteriorGMM}
\end{equation}
where $b[X_{k+1}^{j+}|A_i]\doteq \mathbb{P}(X_{k+1}|\mathcal{H}_{k+1}^-,\gamma=j, A_i, z_{k+1})$ is the posterior of the $j$th GMM component of the propagated belief $b[X_{k+1}^{-}]$, see Eq.~(\ref{eq:PredictedBelief_step1GMM}).

Plugging in Eq.~(\ref{eq:PosteriorGMM}) back into $\prob{X_{k+1}|\mathcal{H}_{k+1}^-,z_{k+1}}$ yields from Eq.~(\ref{eq:Posterior_givenAi}):

\begin{equation}
	b[X_{k+1}]\equiv\prob{X_{k+1}|\mathcal{H}_{k+1}^-,z_{k+1}} =  \sum_{i=1}^{|A_{\mathbb{N}}|} \sum_{j=1}^{M_k}\xi_{k}^j \prob{\event{i} \mid \mathcal{H}_{k+1}^-, z_{k+1}} b[X_{k+1}^{j+}|A_i].
\label{eq:Posterior_BeforePrun}
\end{equation}

The  term, $\prob{\event{i} \mid \his_k, u_k, z_{k+1}}$, is merely the likelihood of \event{i} being actually the one which generated the observation $z_{k+1}$. This term can be evaluated, in a similar fashion to Section \ref{sec:TermA}, accounting for $b[x_{k+1}^{j-}]$ for each considered $j$th component as $\prob{\event{i} \mid \mathcal{H}_{k+1}^-, z_{k+1}} = \int_x \prob{\event{i}, x \mid \mathcal{H}_{k+1}^-, z_{k+1}}$, and applying Bayes' rule yields
\begin{equation}
\tilde{w}_{k+1}^{ij} \!\! \doteq \! \eta' \!\!\! \int_x \!\!  \prob{\! z_{k+1} | A_i, x, \mathcal{H}_{k+1}^-\!} \prob{A_i | \mathcal{H}_{k+1}^-, x \!} b[x_{k+1}^{j-} \!\!=\! x],
\label{eq:WeightGMMcomponent}
\end{equation}
with $\eta' = 1/ \prob{z_{k+1} \mid \mathcal{H}_{k+1}^-}$. Note that for each component $j$, $\sum_i \tilde{w}_{k+1}^{ij}=1$. Finally, we can re-write Eq.~(\ref{eq:Posterior_BeforePrun}) as
\begin{equation}
\prob{X_{k+1}|\mathcal{H}_{k+1}^-,z_{k+1}\!} \!= \!\! \sum_{r=1}^{M_{k+1}} \!\! \xi_{k+1}^r  \mathbb{P}(\!X_{k+1}|\mathcal{H}_{k+1},\gamma=r\!),
\label{eq:Posterior_afterPrune}
\end{equation}
or in short, $b[X_{k+1}] =  \sum_{r=1}^{M_{k+1}}  \xi_{k+1}^r  b[X_{k+1}^{r+}]$, where 
\begin{equation}
\xi_{k+1}^r \doteq \xi_{k+1}^{ij} \equiv \xi_{k}^j \tilde{w}_{k+1}^{ij} \ \ , \ \ b[X_{k+1}^{r+}]\doteq b[X_{k+1}^{j+}|A_i].
\label{eq:WeightGMMcomponent_final}
\end{equation}
As seen, we got a new GMM with $M_{k+1}$ components, where each component $r\in [1,M_{k+1}]$, with appropriate mapping to indices $(i,j)$ from Eq.~(\ref{eq:Posterior_BeforePrun}), is represented by weight $\xi_{k+1}^r$ and posterior conditional belief $b[X_{k+1}^{r+}]$. The latter can be evaluated as the Gaussian $b[X_{k+1}^{r+}] \propto b[X_{k+1}^{j-}] \prob{z_{k+1} \mid x_{k+1}, A_i}=\mathcal{N}(\hat{X}^{r}_{k+1}, \Sigma^{r}_{k+1})$, where the mean $\hat{X}^{r}_{k+1}$ and covariance $\Sigma^{r}_{k+1}$ can be efficiently recovered via MAP inference.

\subsection{Summary thus Far}

To summarize the discussion thus far, we have shown that for the myopic case, the objective function (\ref{eq:cost_function}) can be re-written as
\begin{equation}
J(u_k) = \int_{z_{k+1}} (\sum_i^{|A_{\mathbb{N}}|} w_{k+1}^i) \cdot c\left(\sum_r^{M_{k+1}}  \xi_{k+1}^r   b[X_{k+1}^{r+}] \right).
\label{eq:ObjFuncNew}
\end{equation}

One can observe that according to Eq.~(\ref{eq:Posterior_BeforePrun}), each of the $M_k$ components from the belief at a previous time, is split into $|A_{\mathbb{N}}|$ new components with appropriate weights. This would imply an explosion in the number of components, making the proposed framework hardly applicable. However, in practice, the majority of the weights will be negligible, and therefore can be pruned, while the remaining number of components is denoted by $M_{k+1}$ in Eq.~(\ref{eq:Posterior_afterPrune}). Depending on the scenario and the degree of perceptual aliasing, this can correspond to \emph{full} or \emph{partial} disambiguation.

Having shown incorporating data association within belief space planning leads to Eq.~(\ref{eq:ObjFuncNew}), we now proceed with the exposition of our approach. 

\subsection{Simulating Future Observations $\{z_{k+1}\}$ given $b[X_{k+1}^-]$}
\label{sec:SimObservations}
Calculating the objective function (\ref{eq:ObjFuncNew}) for each candidate action $u_k$ involves considering all possible realizations of $z_{k+1}$. One approach to perform this in practice, is to simulate future observations $\{z_{k+1}\}$ given  propagated GMM belief $b[X_{k+1}^-]$, scenes $\events$ and observation model (\ref{eq:MotionObsModel}). One can then evaluate Eq.~(\ref{eq:ObjFuncNew}) considering all observations in $\{z_{k+1}\}$.

We now briefly describe how this concept can be realised. First, viewpoints $\{x\}$ are sampled from $b[X_{k+1}^-]$. For each viewpoint $x \in \{x\}$, an observed scene $A_i$ is determined according to event likelihood $\prob{\event{i}\mid \mathcal{H}_{k},x}$. Together, $x$ and $A_i$ are then used to generate nominal $\hat{z}=h(x,A_i)$ and noise-corrupted observations $\{z\}$ according to observation model (\ref{eq:MotionObsModel}): $z=h(x,A_i)+v$. The set $\{z_{k+1}\}$ is then the union of all such generated observations $\{z\}$. Note that while generating $\{z_{k+1}\}$, the true association is known (scene $A_i$), it is unknown to our algorithm, i.e.~while evaluating Eq.~(\ref{eq:ObjFuncNew}).
\subsection{Computing Mixture of Posterior Beliefs $\sum_i \tilde{w}_i  b[X_{k+1}^{i+}]$}
As seen from Eq.~(\ref{eq:ObjFuncNew}), reasoning about data association aspects resulted in a mixture of posteriors within the cost $c(.)$, i.e.~$\sum_i \tilde{w}_i  b[X_{k+1}^{i+}]$, for each possible observation $z_{k+1}\in  \{z_{k+1}\}$. In this section we briefly describe how one can actually calculate the corresponding posterior distributions, given some specific observation $z_{k+1}\in \{z_{k+1}\}$. For simplicity, we consider  the belief at planning time $k$ is a Gaussian $b[X_k]=\mathcal{N}(\hat{X}_k, \Sigma_k)$. However, our approach could be applied also to more general cases (e.g.~mixture of Gaussians) with a certain price in terms of computational complexity. Further investigation of these aspects is left to future research.

Under this setting, each of the components $b[X_{k+1}^{i+}]$ in the mixture pdf can be written as $b[X_{k+1}^{i+}] \propto b[X_{k}] \prob{x_{k+1} \mid x_k, u_k} \prob{z_{k+1} \mid x_{k+1}, A_i}$. It is then not difficult to show that the above belief is a Gaussian $b[X_{k+1}^{i+}] = \mathcal{N}(\hat{X}^{i}_{k+1}, \Sigma^{i}_{k+1})$ and to find its first two moments via MAP inference. Obviously, the mixture of posterior beliefs in the cost $c(.)$ from Eq.~(\ref{eq:ObjFuncNew}) is now a mixture of Gaussians:

\begin{eqnarray}
\sum_i \tilde{w}_i  b[X_{k+1}^{i+}] = \sum_i \tilde{w}_i  \mathcal{N}(\hat{X}^{i}_{k+1}, \Sigma^{i}_{k+1}).
\label{eq:mixtureGaussians}
\end{eqnarray}

\subsection{Designing a Specific Cost Function}
\label{sec:design-specific-cost}
 The treatment so far has been agnostic to the structure of the cost function $c(.)$. Recalling Eq.~(\ref{eq:ObjFuncNew}) we see that the belief over which the cost function is defined, is multimodal in general. Standard cost functions in literature, typically include terms such as control usage $c_u$, distance to goal $c_G$ and uncertainty $c_{\Sigma}$, see e.g.~\cite{VanDenBerg12ijrr, Indelman15ijrr}. In our case, however, the specific form of the latter should be re-examined and an additional term quantifying ambiguity level can be introduced. In this section we thus briefly discuss these two terms, starting with the cost over posterior uncertainty.

Since, unlike in usual BSP, the posterior belief in our case is multimodal and represented as mixture of Gaussians $\sum_i \tilde{w}_i  \mathcal{N}(\hat{X}^{i}_{k+1}, \Sigma^{i}_{k+1})$, see Eq.~(\ref{eq:mixtureGaussians}), we could define several different cost structures depending on how we treat the  different modes. Two particular such costs are taking the worst-case covariance among all covariances $\Sigma^{i}_{k+1}$ in the mixture, e.g.~$\Sigma=\max_i \{tr(\Sigma_i)\}$, or to collapse the mixture into a single Gaussian $\mathcal{N}(., \Sigma)$, see e.g.~\cite{BarShalom04book}. In both cases, we can define the cost due to uncertainty as $c_\Sigma = trace(\hat\Sigma)$.

The cost due to ambiguity, $c_w$, should penalise ambiguities such as those arising out of perceptual aliasing. Here, we note that non-negligible weights $w_i$ in Eq.~(\ref{eq:ObjFuncNew}) arise due to perceptual aliasing, 
whereas in case of no aliasing, all but one of these weights are zero. In most severe case of aliasing (all scenes or objects $A_i$ are identical), all of these weights are comparable among each other. Thus we take Kullback-Leibler divergence $KL_u(\{\tilde{w}_i\})$ of these weights $\{\tilde{w}_i\}$ from a uniform distribution to penalise higher aliasing, and define $c_w(\{\tilde{w}_i\})\doteq \frac{1}{ KL_u(\{\tilde{w}_i\})+\epsilon}$, where $\epsilon$ is a small number to avoid division-by-zero in case of extreme perceptual aliasing.
With user-defined weights $M_u, M_G, M_{\Sigma}$ and $M_w$, the overall cost then can be defined as a combination
\begin{equation}
c \doteq M_uc_u + M_Gc_G + M_{\Sigma} c_\Sigma + M_w c_w,
\label{eq:overall_cost}
\end{equation}

\subsection{Formal Algorithm for DA-BSP}
 We now have all the ingredients to present the overall framework of data-association aware belief space planning, calling it DA-BSP for brevity. It is summarised in Algorithm~\ref{alg:bsp-no-da} and briefly described below.

Given a GMM belief $b[X_k]$ and candidate action $u_k$, we first propagate the belief to get $b[X_{k+1}^-]$ and then simulate future observations $\{z_{k+1}\}$ (line \ref{lst:line:sim-msr}). The algorithm then calculates the contribution of each observation $z_{k+1}\in \{z_{k+1}\}$ to the objective function (\ref{eq:ObjFuncNew}). In particular, on lines \ref{lst:line:compute-wt} and \ref{lst:line:compute-wt_accum} we calculate the weights $w_{k+1}^i$ that are used in evaluating the likelihood $w_s$ of obtaining observation $z_{k+1}$. On lines \ref{lst:line:eval_posterior-start}-\ref{lst:line:eval_posterior-end} we compute the posterior belief: this involves updating each $j$th component from the propagated belief $b[X_{k+1}^{j-}]$ with observation $z_{k+1}$, considering each of the possible scenes $A_i$. After pruning (line \ref{lst:line:pruning}), this yields a posterior GMM with $M_{k+1}$ components. We then evaluate the cost $c(.)$ (line \ref{lst:line:evaluate-cost}) and use $w_s$ to update the value of the objective function $J$ with the weighted cost for measurement $z_{k+1}$ (line \ref{lst:line:updateJ}).

\begin{algorithm}
	\scriptsize
	\caption{Data association aware belief-space planning
		\label{alg:bsp-no-da}}
	\begin{algorithmic}[1]
		\Require{Current GMM belief $b[X_k]$ at step-$k$, history $\his_k$, action $u_k$, scenes \events, event likelihood $\prob{\event{i}\mid \mathcal{H}_{k},x}$ for each $A_i\in\events$}
		\Statex
		
		\Let{$b[X_{k+1}^-]$}{$b[X_k] \prob{x_{k+1} \mid x_k,u_k}$} \Comment{Eq.~(\ref{eq:PredictedBelief_step1GMM})}
		\Let{$\{z_{k+1}\}$}{ {\tt SimulateObservations}($b[X_{k+1}^-]$, \events)}\label{lst:line:sim-msr}
		\Let{$J$}{0} 
		\For{$\forall z_{k+1} \in \{z_{k+1}\}$}		
		\Let{$w_s$}{0}
		\For{$i \in [1\dots |\event{}|]$}
		\LineComment{{compute weight}, Eq.~(\ref{eq:wi_def})}
		\Let{$w^i_{k+1}$}{{\tt CalcWeights}($z_{k+1},\prob{\event{i}\mid \mathcal{H}_{k+1}^-,x}, b[X_{k+1}^-]$)}\label{lst:line:compute-wt}
		\Let{$w_s$}{$w_s + w_i$}				\label{lst:line:compute-wt_accum}
		\For{$\forall j \in [1,\ldots,M_k]$}	\label{lst:line:eval_posterior-start}
		\LineComment{{compute weight $\tilde{w}^{ij}_{k+1}$ for each GMM component}, Eq.~(\ref{eq:WeightGMMcomponent})}
		\Let{$\tilde{w}^{ij}_{k+1}$}{{\tt CalcWeights}($z_{k+1},\prob{\event{i}\mid \mathcal{H}_{k+1}^-,x}, b[X_{k+1}^{j-}]$)}
		
		\Let{$\xi_{k+1}^{ij}$}{$\xi_{k}^j \tilde{w}_{k+1}^{ij}$} \Comment{Eq.~(\ref{eq:WeightGMMcomponent_final})}
		\LineComment{Calculate posterior of $b[X_{k+1}^{j-}]$,  given $A_i$}
		\Let{$b[X_{k+1}^{ij+}]$}{ {\tt UpdateBelief}($b[X_{k+1}^{j-}], z_{k+1}, A_i$)}\label{lst:line:conditional-posterior}
		\EndFor \label{lst:line:eval_posterior-end}
		%
		%
		%
		\EndFor
		\State{Prune components with weights $\xi_{k+1}^{ij}$ below a threshold} \label{lst:line:pruning}
		\State{{Construct $b[X_{k+1}^{+}]$ from the remaining $M_{k+1}$ components} via Eq.~(\ref{eq:Posterior_afterPrune})}
		\Let{$c$}{ {\tt CalcCost}($b[X_{k+1}^{+}]$)} \Comment{Eq.~\ref{eq:overall_cost} }\label{lst:line:evaluate-cost}
		\Let{$J$}{$J+w_s\cdot c$} \label{lst:line:updateJ}
		\EndFor
		\State \Return{$J$}
	\end{algorithmic}
\end{algorithm}

\section{Experimental results}
\label{sec:results}


\subsection{An Abstract Example for DA-BSP}
Consider the problem of robotic manipulation of objects in the kitchen. For simplicity, let us abstract it to a simpler domain of three objects, $|\events| = 3$. We consider a single step control at  time step $k$, from a given belief $b[X_k]$, as well as that of one step ahead $b[X_{k+1}^-]$, and assume the following motion and observation models $f$ and $h$ 

\begin{align}
\label{eq:models}
\begin{split}
 f(x,u) = \left( \begin{array}{cc} 1 & 0\\0 & 1 \end{array}\right) \cdot x + d \; \bigg\{  \begin{array}{cr} [0, 1]^T & \text{if ${u}$ = \emph{up}} \\ \relax [1, 0]^T & \text{if $u$ = \emph{right}} \\ \end{array},
\\
 h(x, A_i) = h_i(x) = \left( \begin{array}{cc} 1 & 0\\0 & 1 \end{array}\right) \cdot (x - x_i) + s_i.
\end{split}
\end{align}
where observations as well as the shift $s_i$ is in an object-centric frame, with $x_i$ representing location of $A_i$. Intuitively, $s_i$ is a simple mechanism to model perceptual aliasing between objects; e.g., identical objects $A_i$ would have the same $s_i$. Figure  \ref{fig:obs_generative} illustrates the process of simulating future observations $\{z_{k+1}\}$ for ${u}_k$ = \emph{up}, considering unique and perceptually aliased scenes (Figures~\ref{fig:obs_no_alias}-\ref{fig:obs_alias}). In particular, a sampled pose $x^{tr}$ used to generate an observation $z_{k+1}\in \{z_{k+1}\}$ is shown in Figure~\ref{fig:gen_xk1}.

Figure \ref{fig:toy-posteriors} demonstrates key aspects in our approach, considering each time a single observation $z_{k+1}$. Our approach reasons about data association and hence we consider each $z_{k+1}$ could have been generated by one of the 3 objects; each such association would fetch us a conditional posterior belief $b[X^{i+}_{k+1}]$ as  denoted by small ellipses.  Finally, we compute the total cost according to  Algorithm~\ref{alg:bsp-no-da}. 

\newcommand{\myfig}[2]{\includegraphics[trim=0.5cm 6cm 0.5cm 6cm,scale=#1]{#2}}
\renewcommand{\myfig}[2]{\includegraphics[scale=#1]{#2}}

\begin{figure*}
\centering
\subfloat[Sampled viewpoints]{
\myfig{0.25}{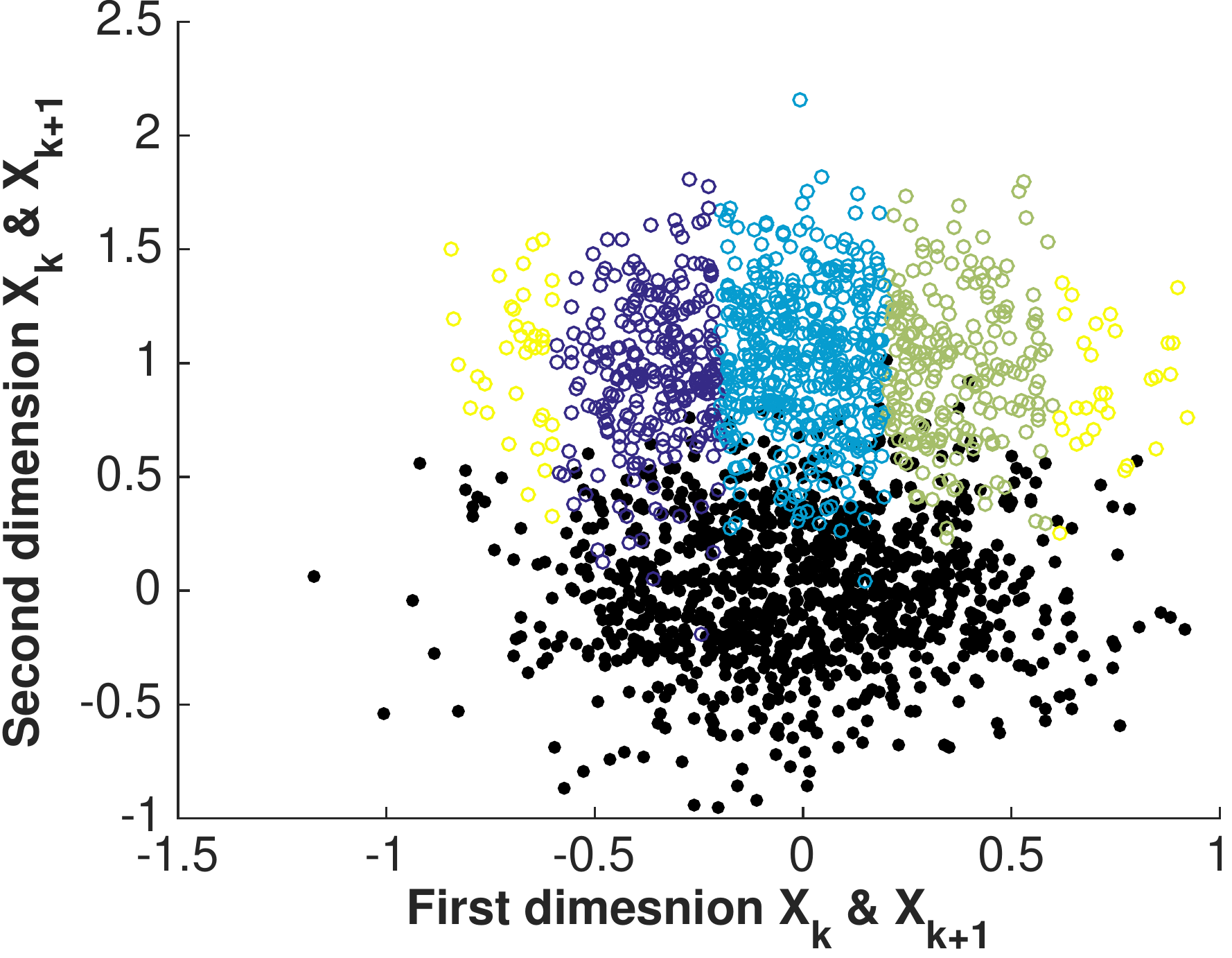} \label{fig:gen_x}}\hspace{0.1\linewidth}
\subfloat[Event likelihood $\prob{A_i|x,\mathcal{H}}$ $\forall i$]{
\includegraphics[trim=0.5cm 6cm 0.5cm 6cm,scale=0.25]{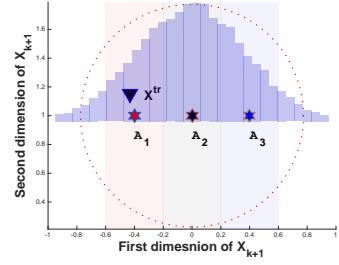} \label{fig:gen_xk1}}\\
\subfloat[No aliasing, \alias{\Phi}]{
\myfig{0.25}{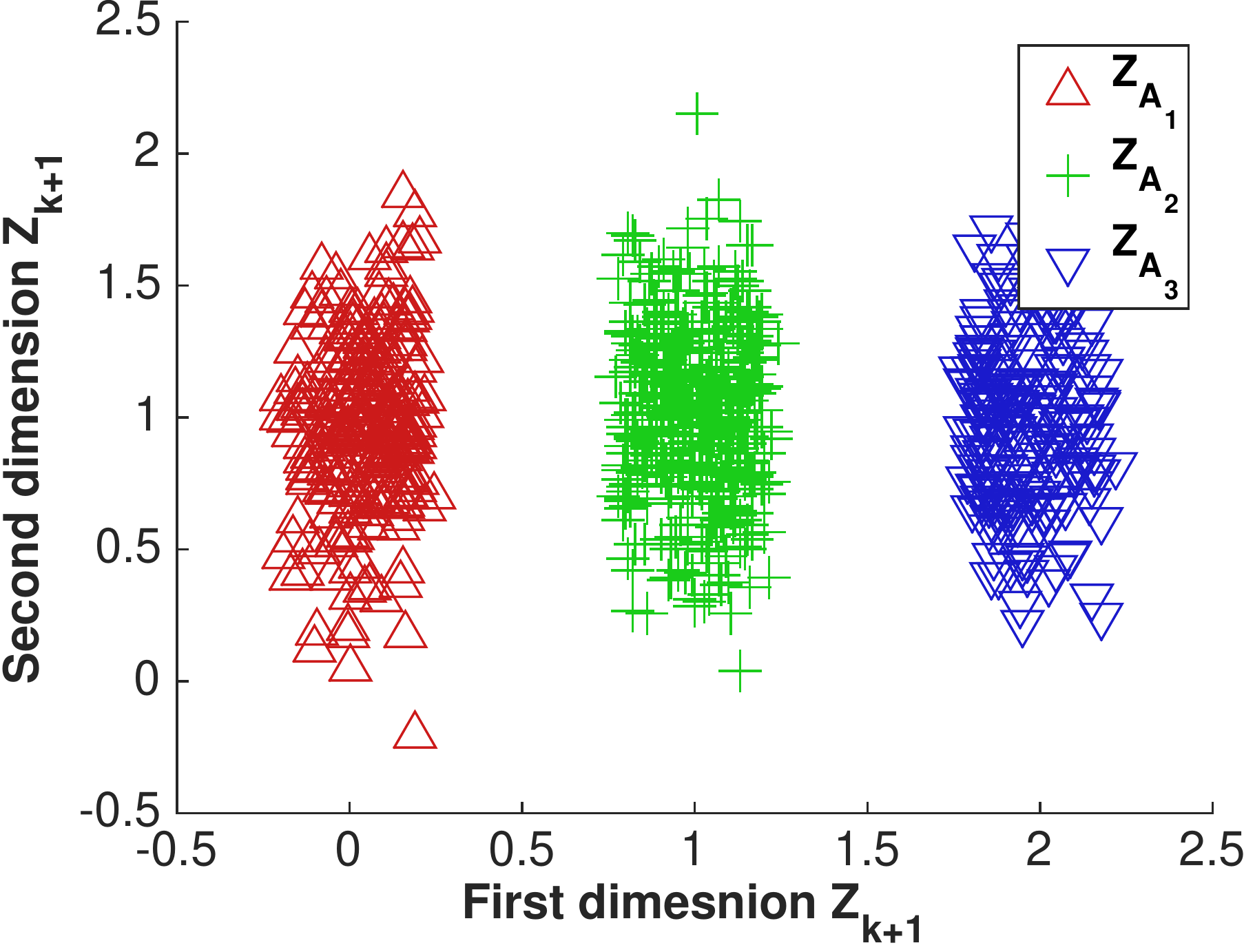} \label{fig:obs_no_alias}}\hspace{0.1\linewidth}
\subfloat[$\alias{\event{1},\event{3}}$]{
\myfig{0.25}{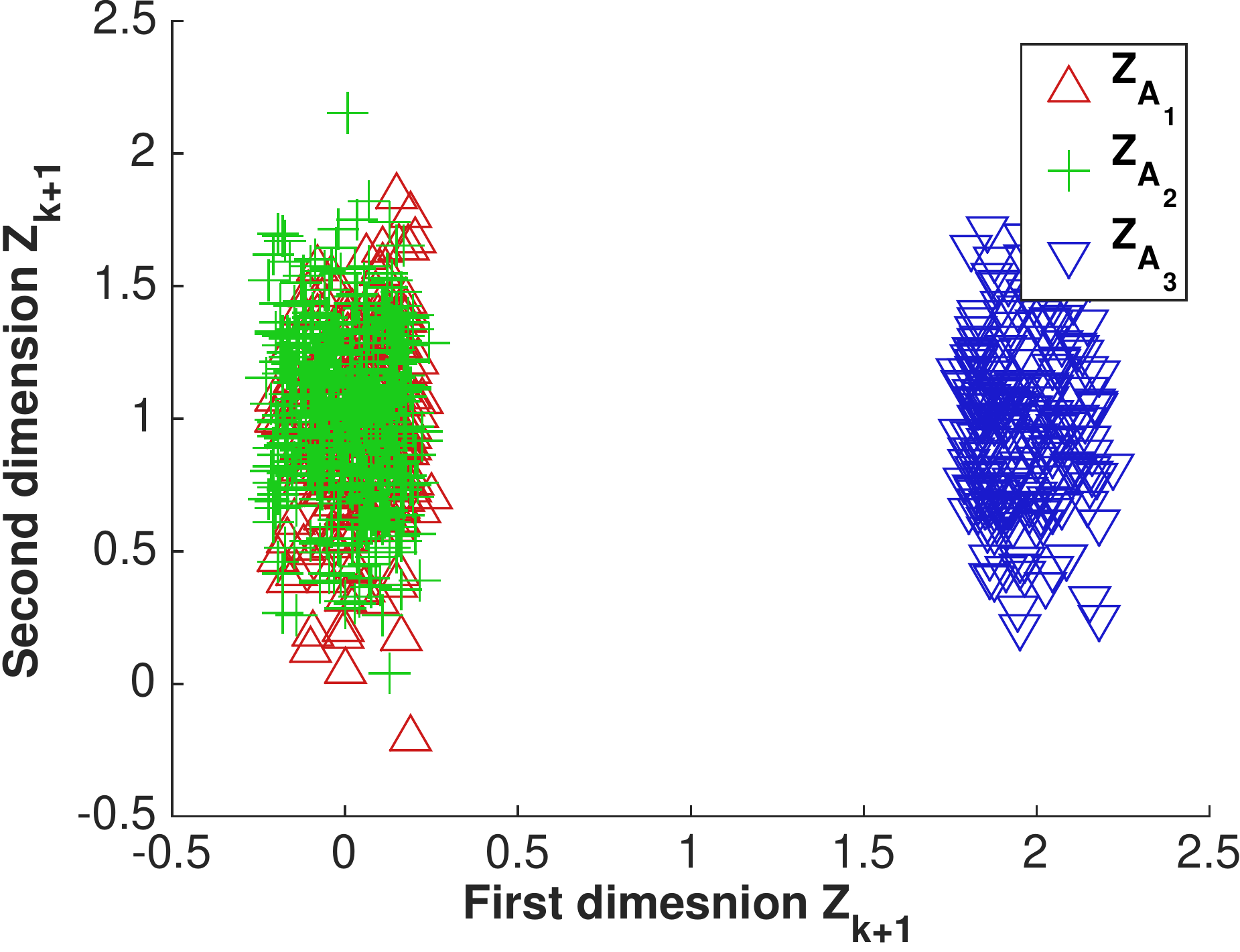} 
\label{fig:obs_alias}}
\caption{ Pose and observation space. (a) black-colored samples $\{x_{k}\}$ are drawn from  $b[X_k]\doteq{\cal N}([0,0]^T, \Sigma_k)$, from which, given control $u_k$, samples $\{x_{x+1}\}$ are computed,  colored according to different scenes \event{i} being observed, and used to generate observations $\{z_{k+1}\}$. (b) Stripes represent locations from which each scene $A_i$ is observable, histogram represents distribution of $\{x_{x+1}\}$, which corresponds to $b[X^-_{k+1}]$. (c)-(d) distributions of $\{z_{k+1}\}$ without aliasing and when  $\alias{\event{1},\event{3}}$.
}
\label{fig:obs_generative}
\end{figure*}
%
Figures~\ref{fig:posterior_NN_c}-\ref{fig:posterior_AA_c} denote the situation when the true pose $x^{tr}$ is close to center and observe \event{2}, while in Figures~\ref{fig:posterior_NN_s}-\ref{fig:posterior_AA_s} it is at the left side and observe \event{1}. Different degrees of aliasing are considered. Both weights $w_i$ and $\tilde w_i$ are shown in the inset histograms. Note that the unnormalised weight $w_i$ is higher when the object is at the centre, because the overall likelihood of the observation is higher. Also, with no aliasing, for any other scene \event{j} than the true one, the normalised weight $w_j$ is small irrespective of where $x^{tr}$ is.
 In other words, weights are also related to how likely the objects are to be the causes behind an observation; in case of no aliasing, this can be negligibly small. This is crucial since it implies that DA-BSP in practical applications with infrequent aliasing, would not require any significant additional computational effort w.r.t.~usual BSP. 

\renewcommand{\myfig}[1]{\includegraphics[trim=0.5cm 6cm 0.5cm 6cm,scale=0.19]{#1}}
\begin{figure*}
\centering
\subfloat[\scriptsize No aliasing i.e., $\alias{\Phi}$, $A^{tr}\!\!=\!\!A_2$]{
\myfig{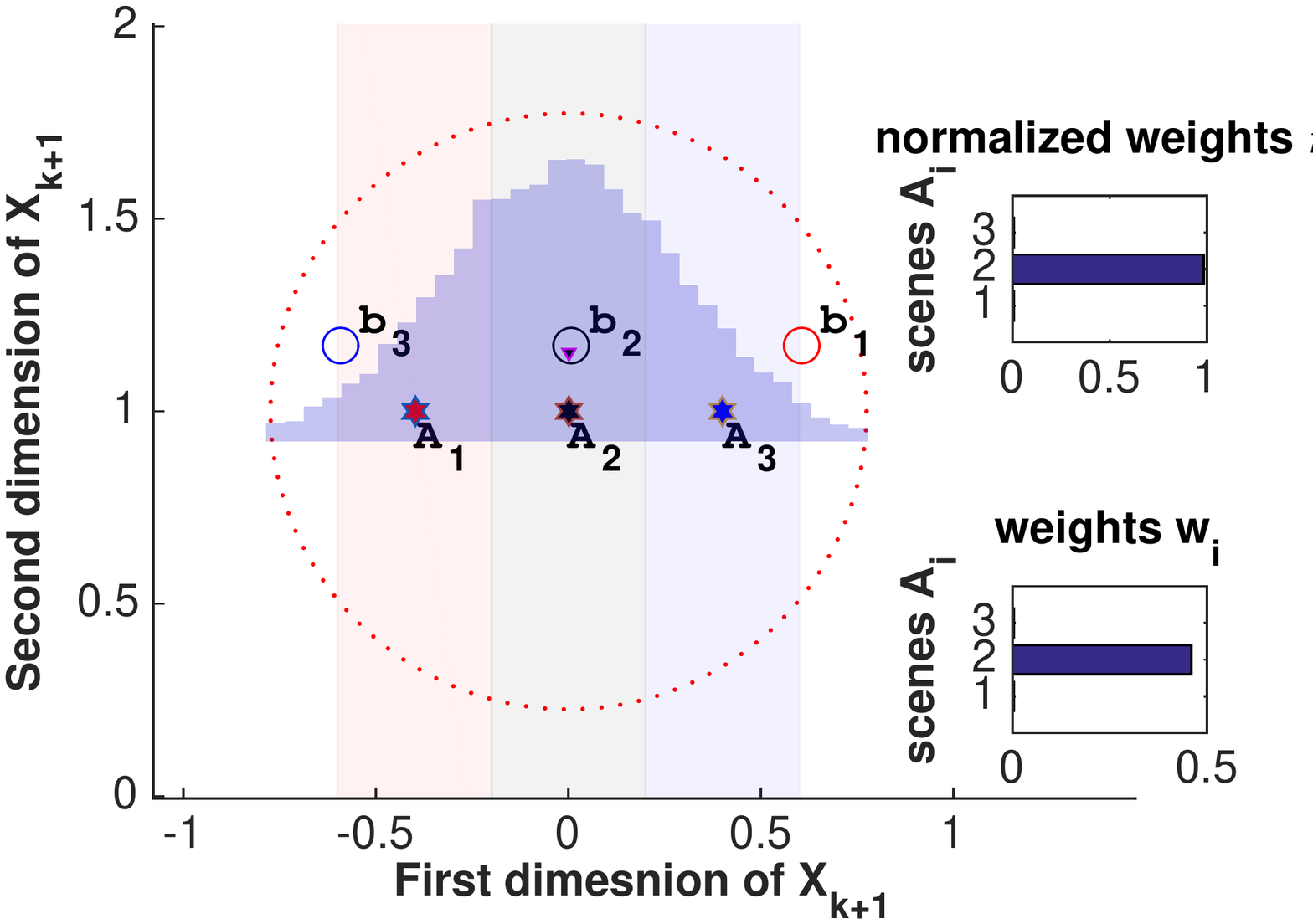}\label{fig:posterior_NN_c}}
\subfloat[\scriptsize $\alias{\event{1},\event{2}}$, $A^{tr}\!\!=\!\!A_2$]{
\myfig{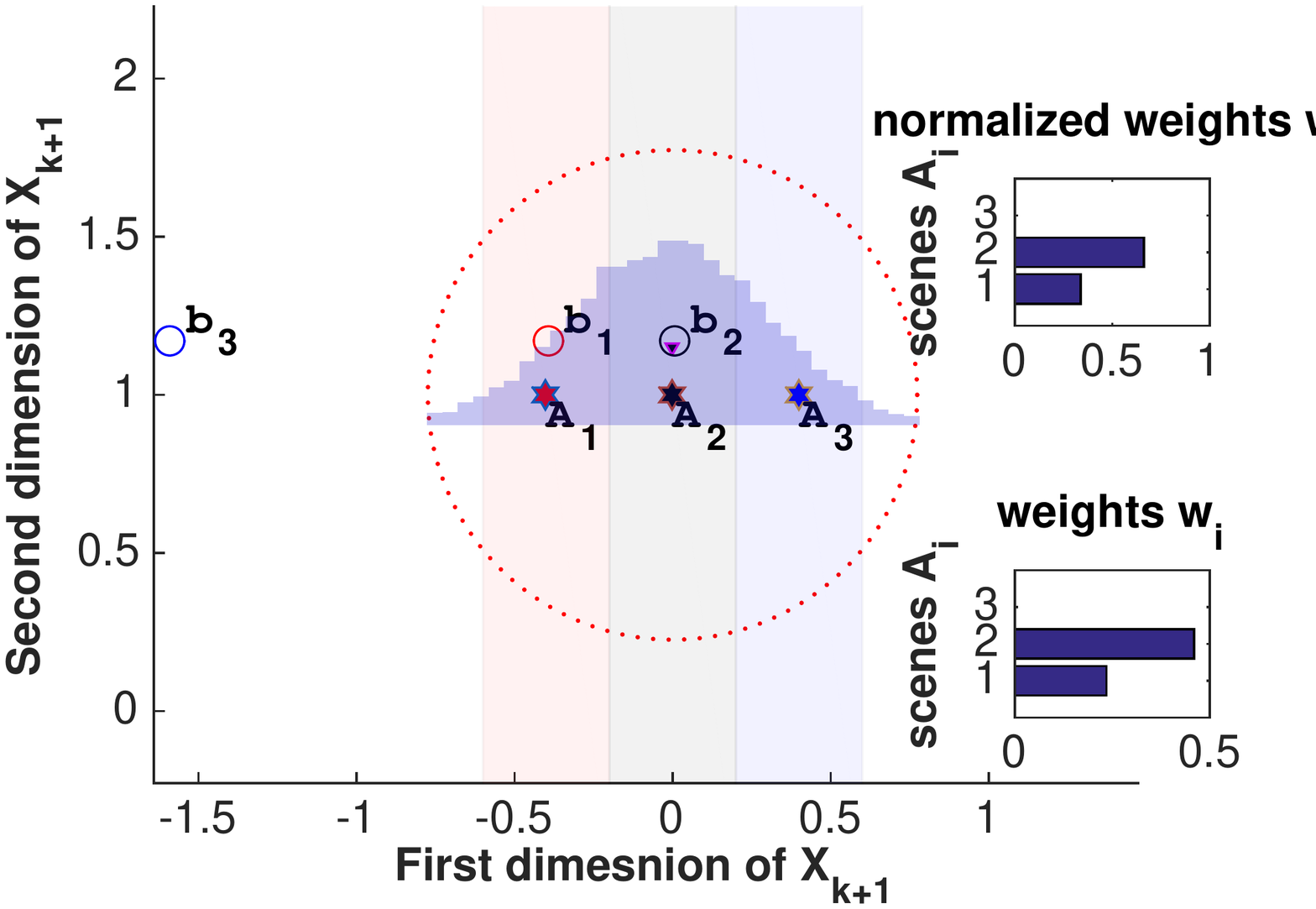}\label{fig:posterior_TA1_c}}
\subfloat[\scriptsize $\alias{\event{1},\event{3}}$, $A^{tr}\!\!=\!\!A_2$]{
\myfig{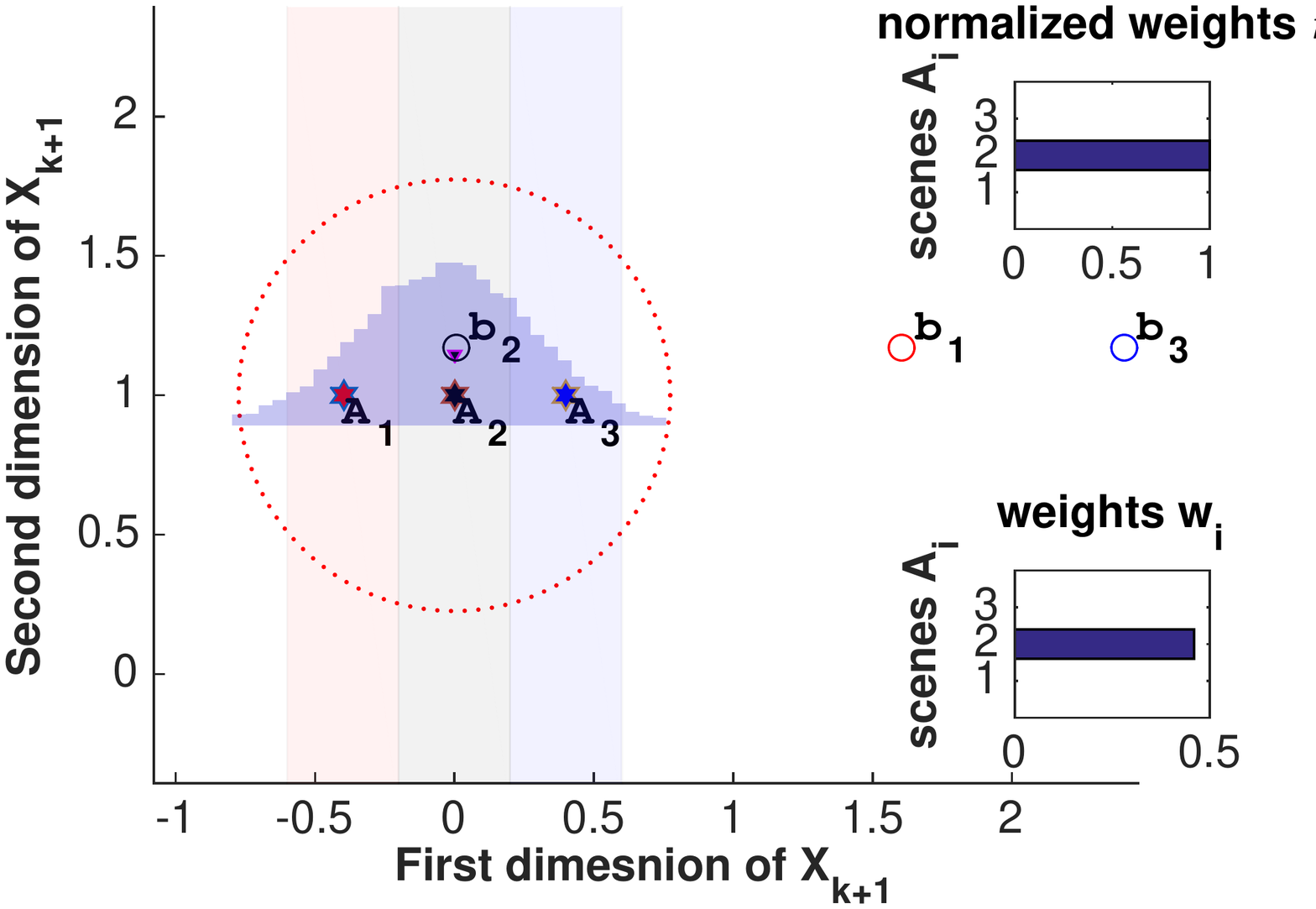}\label{fig:posterior_TA2_c}}\\
\subfloat[$\alias{\event{1},\event{2}, \event{3}}$, $A^{tr}\!\!=\!\!A_2$]{
\myfig{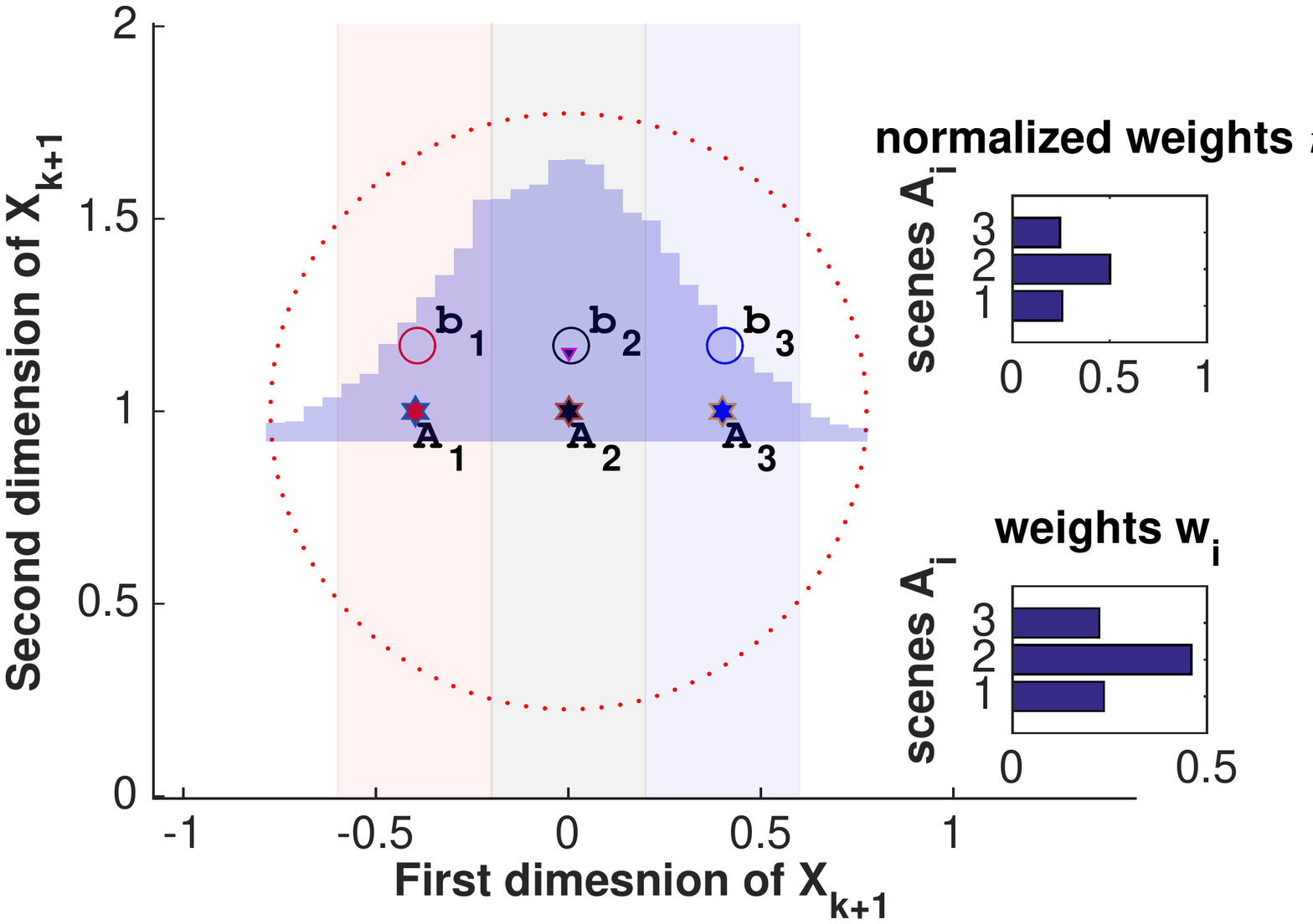}\label{fig:posterior_AA_c}}
\subfloat[\scriptsize $\alias{\Phi}$, $A^{tr}=A_1$]{
\myfig{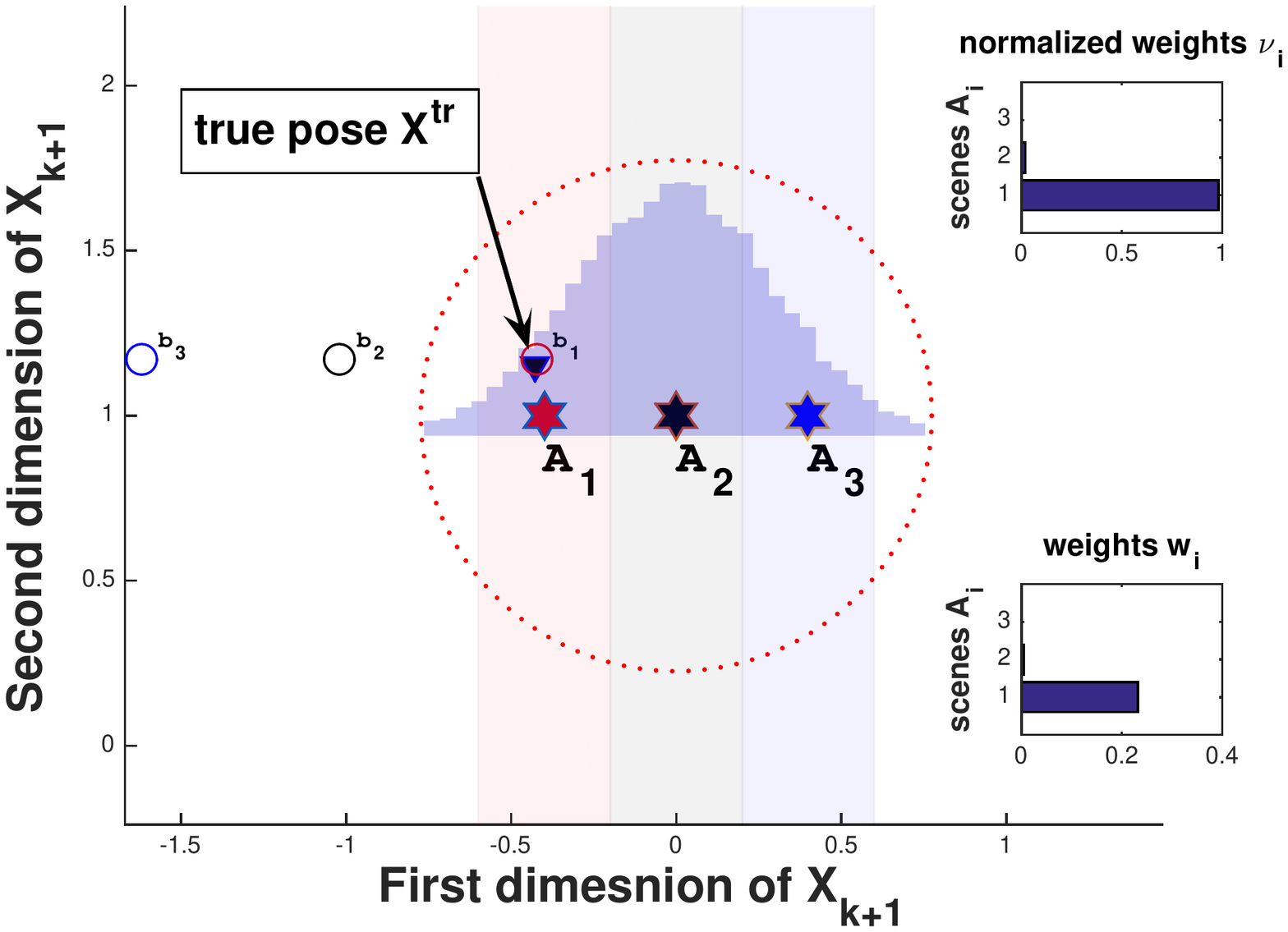}\label{fig:posterior_NN_s}}
\subfloat[\scriptsize $\alias{\event{1},\event{2}}$, $A^{tr}\!\!=\!\!A_1$]{
\myfig{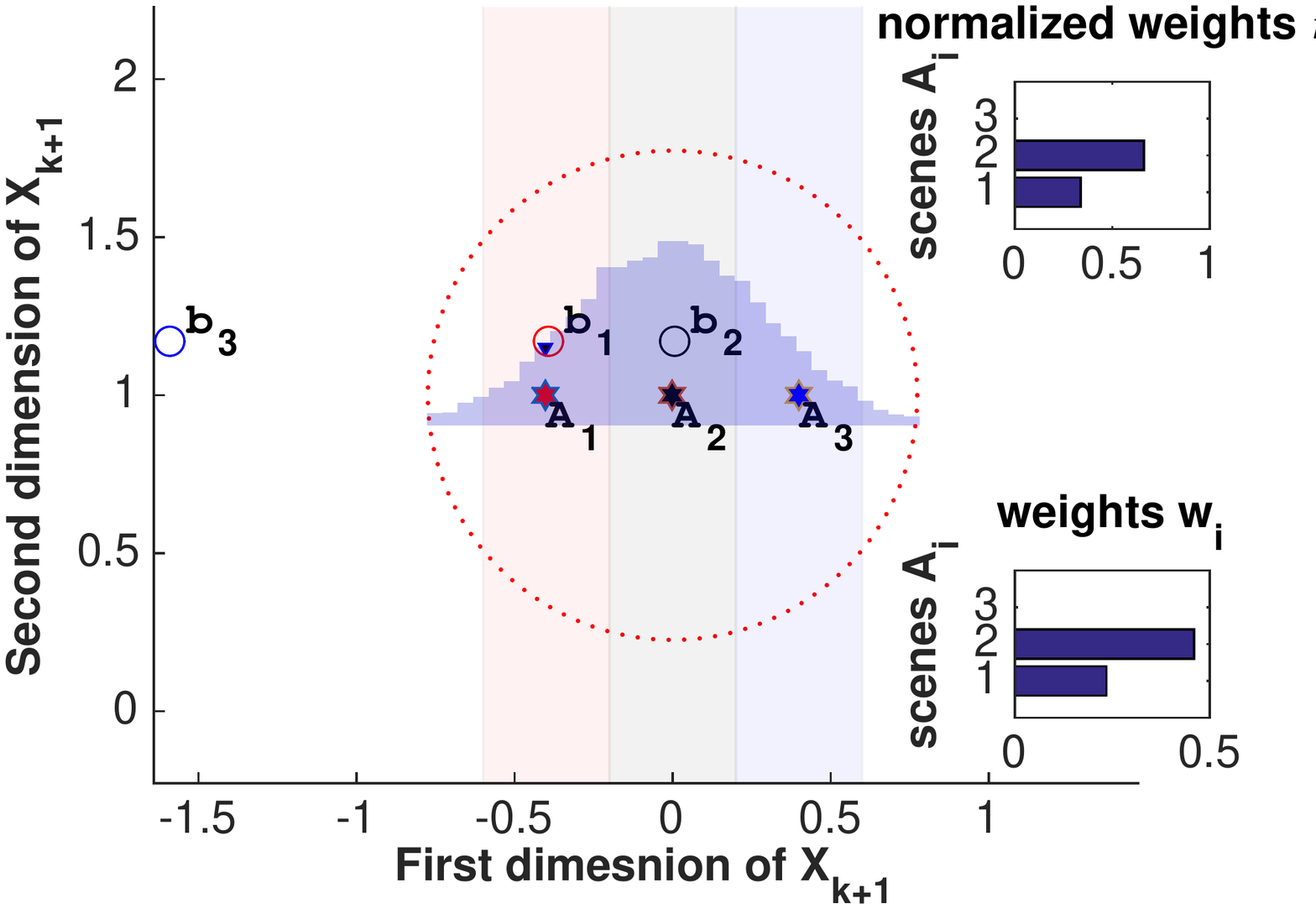}\label{fig:posterior_TA1_s}}\\
\subfloat[\scriptsize $\alias{\event{1}, \event{3}}$, $A^{tr}\!\!=\!\!A_1$]{
\myfig{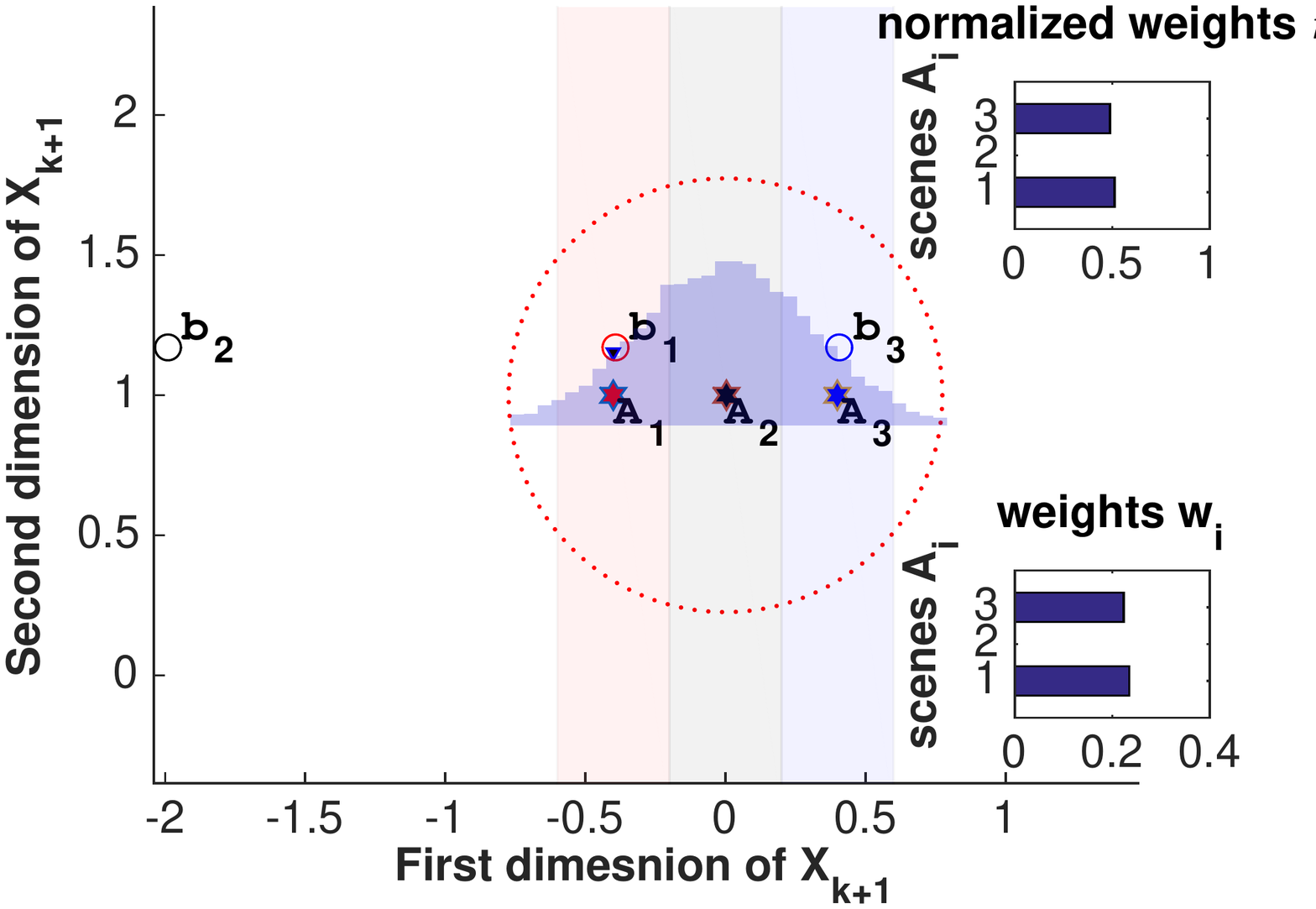}\label{fig:posterior_TA2_s}}
\subfloat[\scriptsize $\alias{\event{1},\event{2}, \event{3}}$, $A^{tr}\!\!=\!\!A_1$]{
\myfig{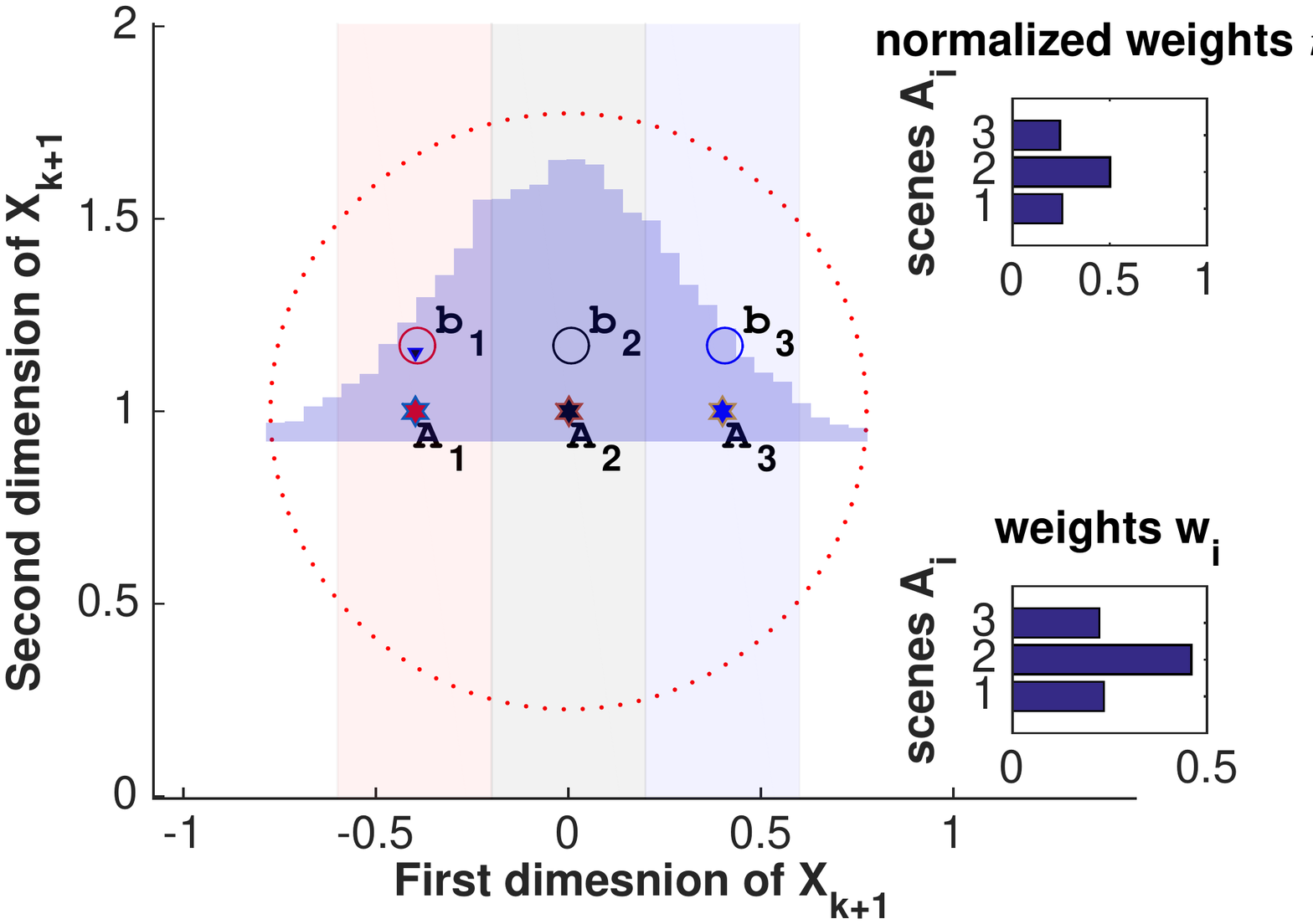}\label{fig:posterior_AA_s}}
\caption{ DA-BSP for a single observation $z_{k+1}$. Red-dotted ellipse denotes $b[X^-_{k+1}]$, while the true pose that generated $z_{k+1}$ is shown by inverted triangle. Smaller ellipses are the posterior beliefs $b[X^{i+}_{k+1}]$. \textbf{(a-d)} $x^{tr}$ is near center, observing $\event{2}$; \textbf{(e-f)} $x^{tr}$ is on the left, observing $\event{1}$. Weights $w_i$ and $\tilde{w}_i$, corresponding to each scene $\event{i}$ are shown in the inset bar-graphs.}
\label{fig:toy-posteriors}
\end{figure*}

Figures~\ref{fig:posterior_TA1_c}-\ref{fig:posterior_AA_c} depict $\alias{\event{1},\event{2}}$, $\alias{\event{1},\event{3}}$ and $\alias{\event{1},\event{2}, \event{3}}$. When $\alias{\event{1},\event{3}}$, the weights $w_i$ are similar, and indeed our cost $c_w$ of weights (in Eq.~(\ref{eq:overall_cost})) is high. For similar uncertainty in pose, this cost would remain constant. Hence, in the presence of identical objects placed similarly within the current belief, optimization of general cost function would be guided towards \emph{active localization}. 
On the other hand, if one object $j$ lies closer to the current nominal pose, it will have slightly higher $w_j$. In case $\alias{\event{1},\event{2}, \event{3}}$, i.e.~\emph{all} objects are identical, the weights $w_i$ are simply an indication of the prior. This is reasonable since in such a case, considering different data association does not yield any new information. 

Finally, in Table~\ref{fig:eval_table}, we present the numerical analysis of cost computation (see Eq.~(\ref{eq:overall_cost})) of these configurations, as well as a metric $\{ \epsilon_{BSP}, \epsilon_{DA} \}$ quantifying estimation error, defined over incorrect (w.r.t ground-truth) associations through random sampling of various modes. Intuitively $\epsilon_{BSP} \text{ and } \epsilon_{DA}$ evaluate how \emph{good} the posterior mean is w.r.t. ground-truth $x^{tr}$ for usual BSP and DA-BSP respectively (lower is better). Recall that unlike action $u_1$, action $u_2$ leads to fully unambiguous observations, around most-likely value (see Fig.~\ref{fig:toy-posteriors}) and consequently, $\epsilon_{BSP} \simeq \epsilon_{DA}$.

\begin{table}[ht] 
\centering
\begin{threeparttable}
\caption{Evaluating different cost functions for various configurations (see Fig.~\ref{fig:toy-posteriors})}
\label{fig:eval_table}
\scalebox{1}{
\resizebox{\columnwidth}{!}{%
\begin{tabular}{ cc|cccc|cc|r  }
\hline
\multicolumn{2}{c|}{\multirow{ 2}{*}{config}} & \multicolumn{4}{c|}{cost} & \multicolumn{2}{c|}{est. err.} & ${\cal U}$\\
\cline{3-9}
\multicolumn{2}{c|}{} & worst & max-wt. & $KL_u$ & mode & $\epsilon_{BSP}$ & $\epsilon_{DA}$ & $u_1/u_2$\\
\cline{1-9}

\multirow{ 8}{*}{}  & $\alias{\Phi}$ &  0.0977 &  0.0977 &  0.3496 &  1.1000 &  1.0851 &  0.1717 &  $u_1$ \\ 
 & \alias{\Phi}  &  0.1009 &  0.1009 &  -na- &  1.0000 &  0.3461 &  0.3999 &  $u_2$ \\ 
 & \alias{\event{1},\event{2}, \event{3}}  &  0.0508 &  0.0508 &  0.5072 &  3.0000 &  1.1654 &  0.4772 &  $u_1$ \\ 
 & \alias{\event{1},\event{2}, \event{3}}  &  0.1009 &  0.1009 &  -na- &  1.0000 &  0.3832 &  0.3990 &  $u_2$ \\ 
 & \alias{\event{1},\event{2}}  &  0.0833 &  0.0833 &  0.3757 &  1.5500 &  1.2197 &  0.2114 &  $u_1$ \\ 
 & \alias{\event{1},\event{2}}  &  0.1009 &  0.1009 &  -na- &  1.0000 &  0.3912 &  0.3992 &  $u_2$ \\ 
 & \alias{\event{1}, \event{3}}  &  0.0849 &  0.0849 &  0.3649 &  1.4000 &  1.0552 &  0.4197 &  $u_1$ \\ 
 & \alias{\event{1},\event{3}} &  0.1009 &  0.1009 &  -na- &  1.0000 &  0.4101 &  0.3940 &  $u_2$  \\
\hline
\end{tabular}}}

\end{threeparttable}
\end{table}

\vspace{-5pt}

\newcommand{\cnf}{\texttt}

\subsection{Gazebo World}
 To demonstrate generality of DA-BSP, we compared (in simulation) it with current state of the art (denoted as \cnf{BSP}) and the approach proposed in \cite{Agarwal15arxiv}. For the latter case (see Fig.~\ref{fig:counter-example}), we have a simulated environment of rectangular corridors with shelfs($s_i$)  and elevator($e$), where a  pioneer robot has a non-Gaussian belief prior (shown with $p_1$, $p_2$); as it can be in either of the two corridors, with localization as its objective. We evaluate our algorithm for inference (\cnf{infer}) as well as active planning (\cnf{plan}) (see Table \ref{fig:eval_da_bsp}). The absence of pose uncertainty, which is the case in \cite{Agarwal15arxiv}, would lead us to a wrong inference and estimate the robot to be at pose $p_1$ (corridor 1) initially, whereas  our approach which considers pose uncertainty will lead to correct inference with high weight for being in corridor 2 (Fig.~\ref{fig:counter-example}). 

\renewcommand{\myfig}[2]{\includegraphics[scale=#1]{#2}}

\begin{figure*}
\centering
\subfloat[Four floors]{
\myfig{0.25}{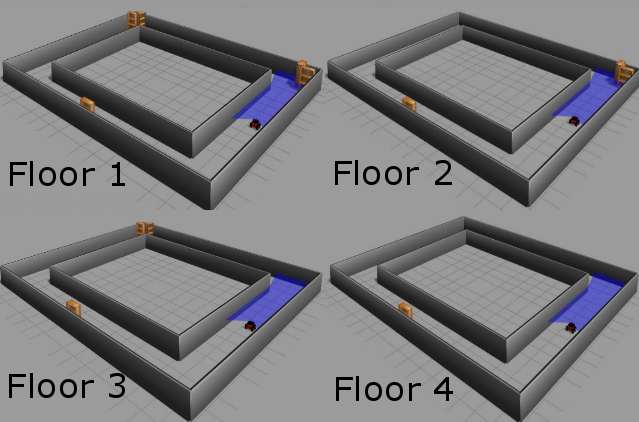} \label{fig:four_floors}}
\subfloat[Counter-example]{
\myfig{0.135}{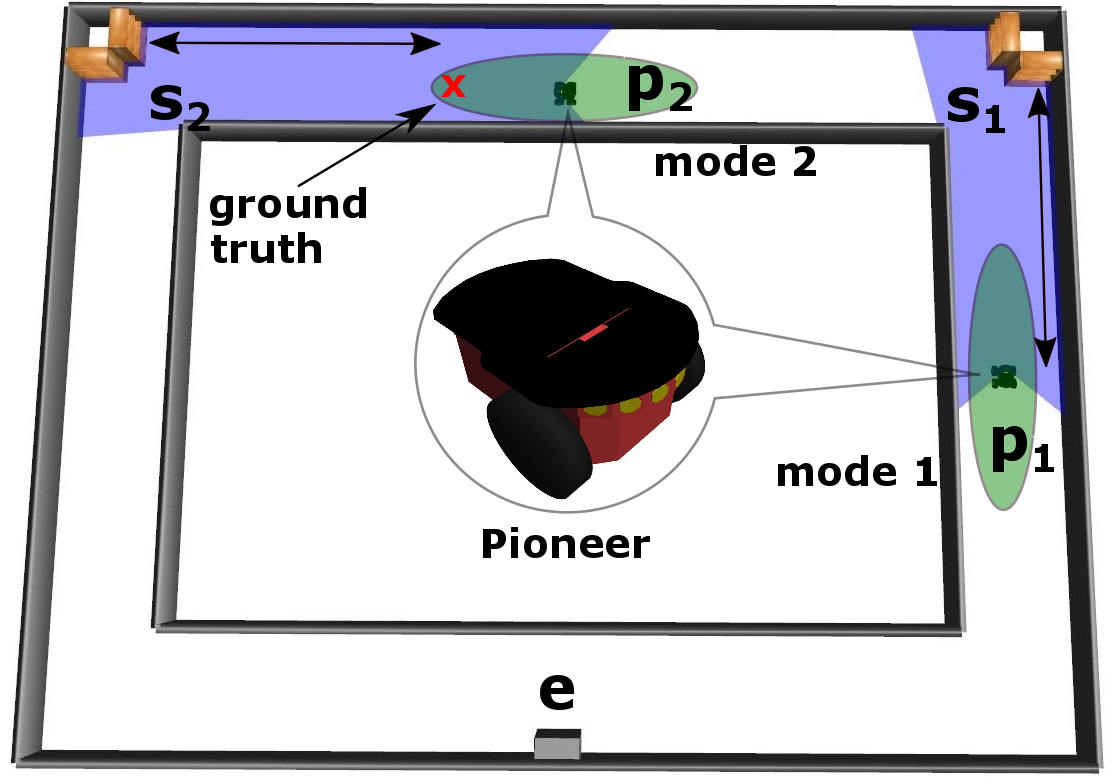} \label{fig:counter-example}}
\caption{ Using Pioneer robot in Gazebo simulation. (a) four-floor aliased world. (b) counter-example for hypothesis reduction in absence of pose-uncertainty in prior. The inference incorrectly deduces that the robot is in mode 1.}
\end{figure*}

To compare with the current state of the art, we consider another scenario with 4 floors. Floor 1 has the same configuration as in Fig.~\ref{fig:counter-example}, floor 2 and floor 3 have the left and the right shelves (w.r.t floor 1) removed respectively and floor 4 has no shelves at all. We use the metric $\eta_{da}$ as one of the possible ways to quantify data association performance by computing the probability of picking the right mode in the posterior GMM (which corresponds to ground-truth position of robot). Thus, for two equally weighted GMM components, $\eta_{da}=0.5\, \text{or} \,0$ if the correct component is one of them or if it has been pruned. The number of modes in the posterior is also indicated (for planning, the total number of modes for \emph{all} the considered future observations is shown). As seen, many of the data associations considered within BSP are wrong ($\eta_{da}=0.29$), while also in inference an incorrect association is made ($\eta_{da}=0$). This can lead to (possibly catastrophic) mission failure; in this case, failure to associate to the correct corridor. We also show a counter-example (Fig.~\ref{fig:counter-example})) for \cite{Agarwal15arxiv} ($\eta_{da}=0$) since that approach does not model uncertainty within each of the GMM components. In contrast, DA-BSP outperforms both approaches within planning and inference.

\renewcommand{\myfig}[2]{\includegraphics[scale=#1]{#2}}

\begin{table}[ht] 

 \begin{threeparttable}
 \caption{Evaluating DA-BSP}
\label{fig:eval_da_bsp}
				\scriptsize
				\begin{tabular}{ lrr|cc|c|c  }
					\hline
					\multicolumn{3}{c|}{\multirow{ 2}{*}{config}} & \multicolumn{2}{c|}{cost} & \multicolumn{2}{c}{metrics}\\
					\cline{4-7}
					\multicolumn{3}{c|}{} & $KL_u$ & Worst-Cov & modes & $\eta_{da}$\\
					\cline{1-7}
					\multirow{ 6}{*}{\rotatebox[origin=c]{90}{\textbf{compare}}} & \multirow{ 2}{*}{\cnf{DA-BSP}}  & \cnf{plan} & 2.60 &	5.48 &	21 &	\textbf{0.41}\\ 
					& & \cnf{infer} & 8.14 &	5.08 &	4 &	\textbf{0.26}\\
					& \multirow{ 2}{*}{\cnf{BSP}}  & \cnf{plan} & -8.67 &	5.36 &	13 &	0.29\\ 
					& & \cnf{infer} & -4.35 &	2.95 &	2 &	0\\ 
					& \multirow{ 2}{*}{\cnf{\cite{Agarwal15arxiv}}}  & \cnf{plan} & -na- &	-na- &	-na- & -na-\\ 
					& & \cnf{infer} & -63.76 &	2.82 &	2 &	0\\ 					
					
					\hline
				\end{tabular}
			\end{threeparttable}
			
\end{table}

In order to evaluate DA-BSP, we consider a simplistic set of actions, namely $\{\cnf{$fwd_1$},\cnf{$fwd_2$},\cnf{$bwd_1$}\}$ for a one-step forward, two-step forward and one-step backward movements, respectively (see Table~\ref{tab:da_bsp_actions}). These actions highlight the challenges of data-association aware planning, even in the context of a simplistic scenario. Note that number of modes (which signifies different associations planner is considering at that step) is significantly higher in the planning, than in the inference. However, when there exists a disambiguating action, such as \cnf{$fwd_2$} is, the planner is able to associate to the correct association all the time (demonstrated by $\eta_{da} = 1$).

\begin{table}[ht] 

 \begin{threeparttable}
 \caption{DA-BSP for candidate actions}
\label{tab:da_bsp_actions}
				\scriptsize
				\begin{tabular}{ lrr|cc|c|c  }
					\hline
					\multicolumn{3}{c|}{\multirow{ 2}{*}{config}} & \multicolumn{2}{c|}{cost} & \multicolumn{2}{c}{metrics}\\
					\cline{4-7}
					\multicolumn{3}{c|}{} & $KL_u$ & Worst-Cov & modes & $\eta_{da}$\\
					\cline{1-7}
					\multirow{ 6}{*}{\rotatebox[origin=c]{90}{\textbf{DA-BSP}}} & \multirow{ 2}{*}{\cnf{$bwd_1$}}  & \cnf{plan} & 6571.29& 	 28.74& 	 48& 	 0.08\\ 
					& & \cnf{infer} & 6567.86& 	 30.53& 	 4& 	 0.08\\
					& \multirow{ 2}{*}{\cnf{$fwd_1$}}  & \cnf{plan} & -1160.93& 	 6.22& 	 22& 0.18\\ 
					& & \cnf{infer} & -1300.72& 	 6.98& 	 2& 	 0.16\\ 
					& \multirow{ 2}{*}{\cnf{$fwd_2$}}  & \cnf{plan} & -166.03& 	 0.66& 	 2& 	 1\\ 
					& & \cnf{infer} & -227.03& 	 0.91& 	 1& 	 1\\ 
					
					\hline
				\end{tabular}
			\end{threeparttable}
			
\end{table}

\subsection{Aliased Multi-Floor Environment}
\begin{figure*}
\centering
\subfloat[Real world]{
\myfig{0.35}{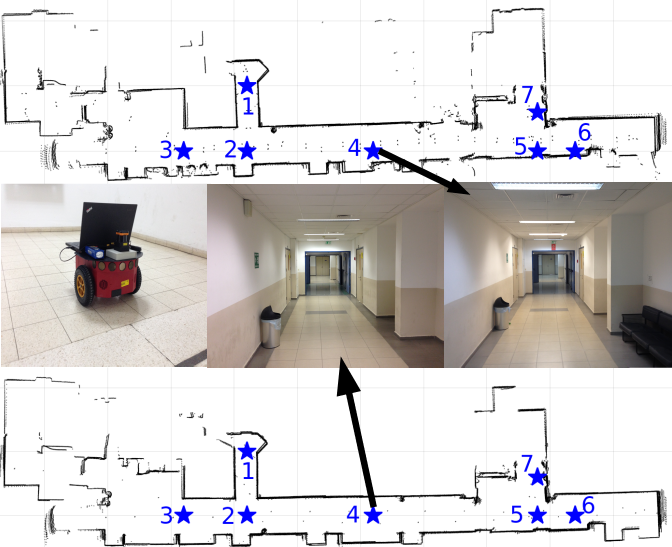} \label{fig:pioneer}}
\subfloat[Partial decision tree]{
\myfig{0.35}{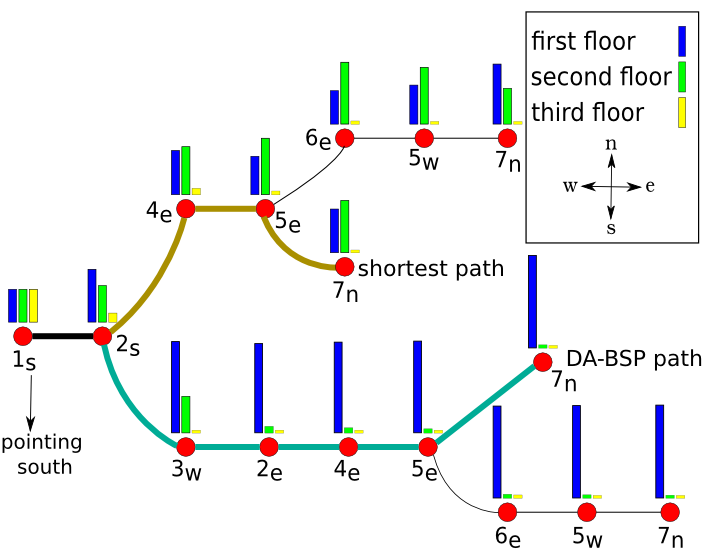} \label{fig:pioneer_tree}}
\caption{ Using Pioneer robot in the real-world. (a) two (of three) severely-aliased floors, and belief space planning for it (b) DA-BSP can plan for  fully disambiguating path (otherwise sub-optimal, due to path-length) while usual BSP with \emph{maximum likelihood} assumption can not (as shown by almost equi-probable modes in the histogram).}
\end{figure*}

 Since, incorporating data-association implies that (at least theoretically) an exponential blow-up of number of unimodal beliefs maintained in the posterior, we therefore evaluated \cnf{DA-BSP} in simple but real-world scene by deploying a real Pioneer robot in a 3-floor aliased environment (see Fig.~\ref{fig:pioneer}), with similar objective of floor and position disambiguation. When not reasoning about perceptual aliasing, the robot takes greedy (w.r.t. control cost and position uncertainty) action and fails to disambiguate, whereas \cnf{DA-BSP} successfully tackles planning with data-association, as shown in Figure \ref{fig:pioneer_tree}. Although not witnessed here, BSP can not guarantee global optimum solution (w.r.t. control and uncertainty cost) and DA-BSP similarly can not provide a global least cost path for full disambiguation - a problem known to be NP-hard~\cite{Dudek98}.

\section{Conclusions}
\label{sec:conclusions}
State-of-the-art belief space planning (BSP) approaches typically consider data association to be given and perfect. However, such an assumption is less appropriate in presence of localisation uncertainty while operating in ambiguous environments, where two scenes could be similar in appearance when observed from appropriate viewpoints. In this work, we developed a data association aware belief space planning (DA-BSP) approach that relaxes the aforementioned assumption. Our framework rigorously incorporates data association aspects within BSP, while considering different sources of uncertainty (uncertainty in robot motion, sensing and possibly in the observed environment). As such, it is capable of better coping with ambiguous, perceptually aliased, situations by appropriately calculating belief evolution and expected cost due to candidate actions, and in particular, could be used for active disambiguation. Thanks to this association being inherent in planning, DA-BSP considers data-association parsimoniously and a simple thresholding is enough for a scalable application of data-association aware belief space planning. We demonstrated key aspects of DA-BSP in abstract example as well as Gazebo simulations. We also applied it on a real-world problem with Pioneer robot lost in a multi-storied building.  

One of the major contributions of this work is in proposing a data-association-aware robust perception in a unified  framework of plan-infer-execute. This is in contrast with passive approaches known in robust perception literature as well as that of multi-hypothesis tracking. Consequently, we are currently looking into extending the approach to non-myopic setting such that the generality of the framework becomes further explicit. Additionally, proving the general theoretical properties of DA-BSP, such as \emph{probabilistic completeness under uncertainty} along the lines proposed by \cite{AghaMohammadi14ijrr}, is another interesting direction of research. Apart from this, evaluating the approach in a more complex real-world scenarios is also an avenue for future research.

\bibliographystyle{plain}
\bibliography{refs}

\end{document}